\documentclass[10pt,twocolumn,letterpaper]{article}

\usepackage{cvpr}
\usepackage{times}
\usepackage{epsfig}
\usepackage{graphicx}
\usepackage{amsmath}
\usepackage{amssymb}
\usepackage{soul}
\usepackage{caption,booktabs,multirow,comment}
\usepackage{subfig}
\usepackage{array}
\newcolumntype{L}[1]{>{\raggedright\let\newline\\\arraybackslash\hspace{0pt}}m{#1}}
\newcolumntype{C}[1]{>{\centering\let\newline\\\arraybackslash\hspace{0pt}}m{#1}}
\newcolumntype{R}[1]{>{\raggedleft\let\newline\\\arraybackslash\hspace{0pt}}m{#1}}

\usepackage[pagebackref=true,breaklinks=true,letterpaper=true,colorlinks,bookmarks=false]{hyperref}

\newcommand{\specialcell}[2][c]{%
  \begin{tabular}[#1]{@{}c@{}}#2\end{tabular}}

 \cvprfinalcopy 

\def\cvprPaperID{256} 

\usepackage[font=small,skip=10pt]{caption}
\ifcvprfinal\fi

\begin{document}

\title{Trajectory Aligned Features For First Person Action Recognition}

\author{ \Large
\quad Suriya Singh \textsuperscript{1}
\quad Chetan Arora \textsuperscript{2}
\quad C. V. Jawahar \textsuperscript{1}\\[6pt]
\normalsize
\textsuperscript{1} IIIT Hyderabad, India
\quad \textsuperscript{2} IIIT Delhi, India
}

\maketitle

\begin{abstract}

Egocentric videos are characterised by their ability to have the first person view. With the popularity of Google Glass and GoPro, use of egocentric videos is on the rise. Recognizing action of the wearer from egocentric videos is an important problem. Unstructured movement of the camera due to natural head motion of the wearer causes sharp changes in the visual field of the egocentric camera causing many standard third person action recognition techniques to perform poorly on such videos. Objects present in the scene and hand gestures of the wearer are the most important cues for first person action recognition but are difficult to segment and recognize in an egocentric video. We propose a novel representation of the first person actions derived from feature trajectories. The features are simple to compute using standard point tracking and does not assume segmentation of hand/objects or recognizing object or hand pose unlike in many previous approaches. We train a bag of words classifier with the proposed features and report a performance improvement of more than $11\%$ on publicly available datasets. Although not designed for the particular case, we show that our technique can also recognize wearer's actions when hands or objects are not visible. 

\end{abstract}

\section{Introduction}
Advances in camera sensors and other related technologies have led to the rise of wearable cameras which are comfortable to use. In the past few years, the use of Google glass \cite{google_glass} and GoPro \cite{GoPro} have become increasingly popular. Such cameras are typically worn on the head or along with the eyeglasses and have the advantage of capturing from a similar point of view as that of the person wearing the camera. We refer to such cameras, with the first person view, as the egocentric cameras.

Excitement of sharing one's actions with the friends and community have made egocentric cameras like GoPro defacto standard in extreme sports. Egocentric cameras can be used to capture daily visual logs for law enforcement officers leading to a significant decrease in complaints against the officers \cite{body_camera_police}. Daily logs from egocentric cameras are also useful in a video sharing application or simply as a memory aid for the wearer. For visually challenged, researchers are trying to augment egocentric videos with meta data such as face, place, text etc. \cite{orcam}. Even for people with regular vision, the promise of giving context aware suggestions is compelling. In spite of their popularity, egocentric videos can be difficult to watch from start to end because of the constant and extreme shake present in such videos due to natural motion of wearer's head.

\begin{figure}[t]
    \centering
    \includegraphics[width=0.48\linewidth]{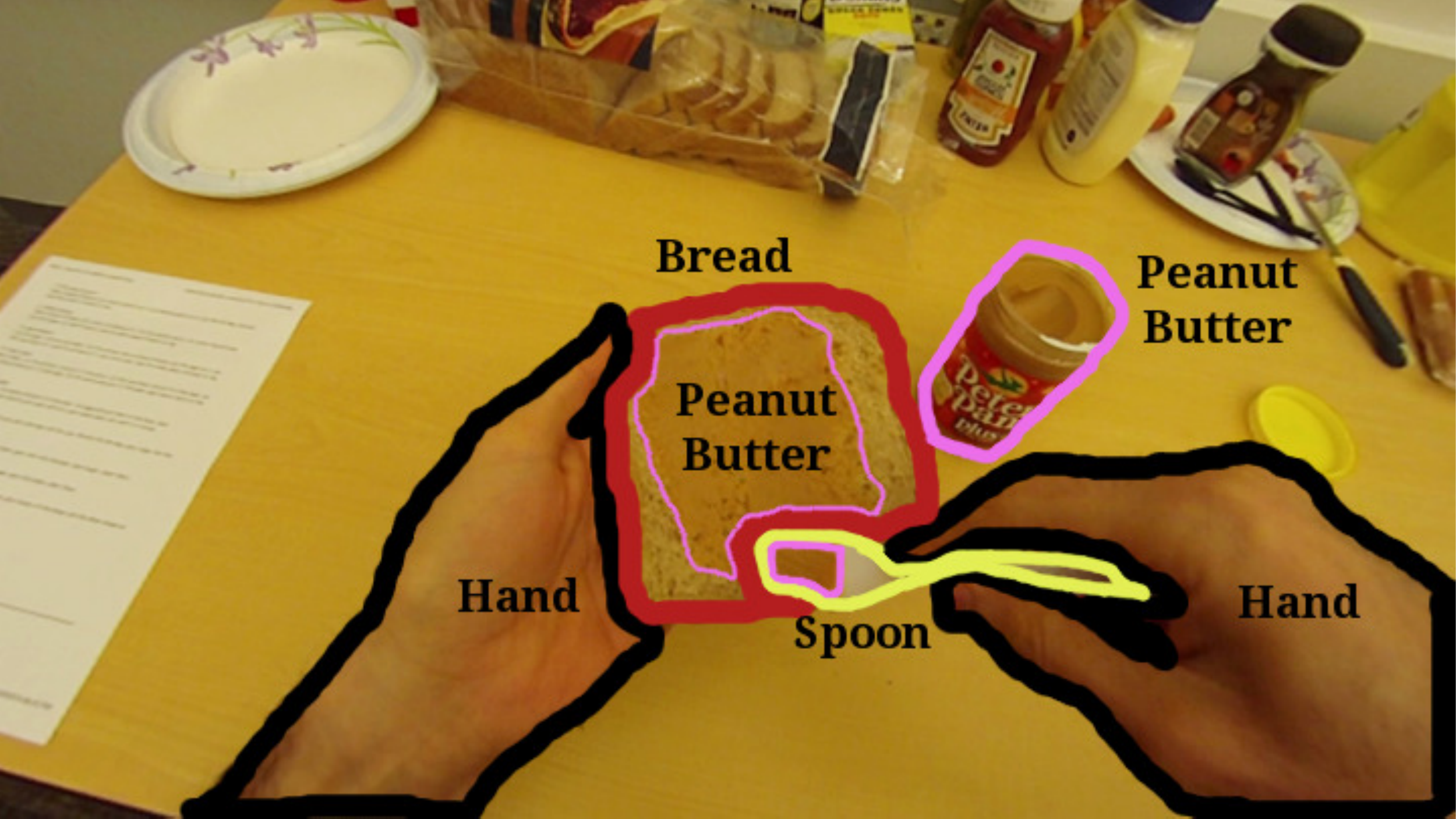} \vspace{0.05cm} \includegraphics[width=0.48\linewidth]{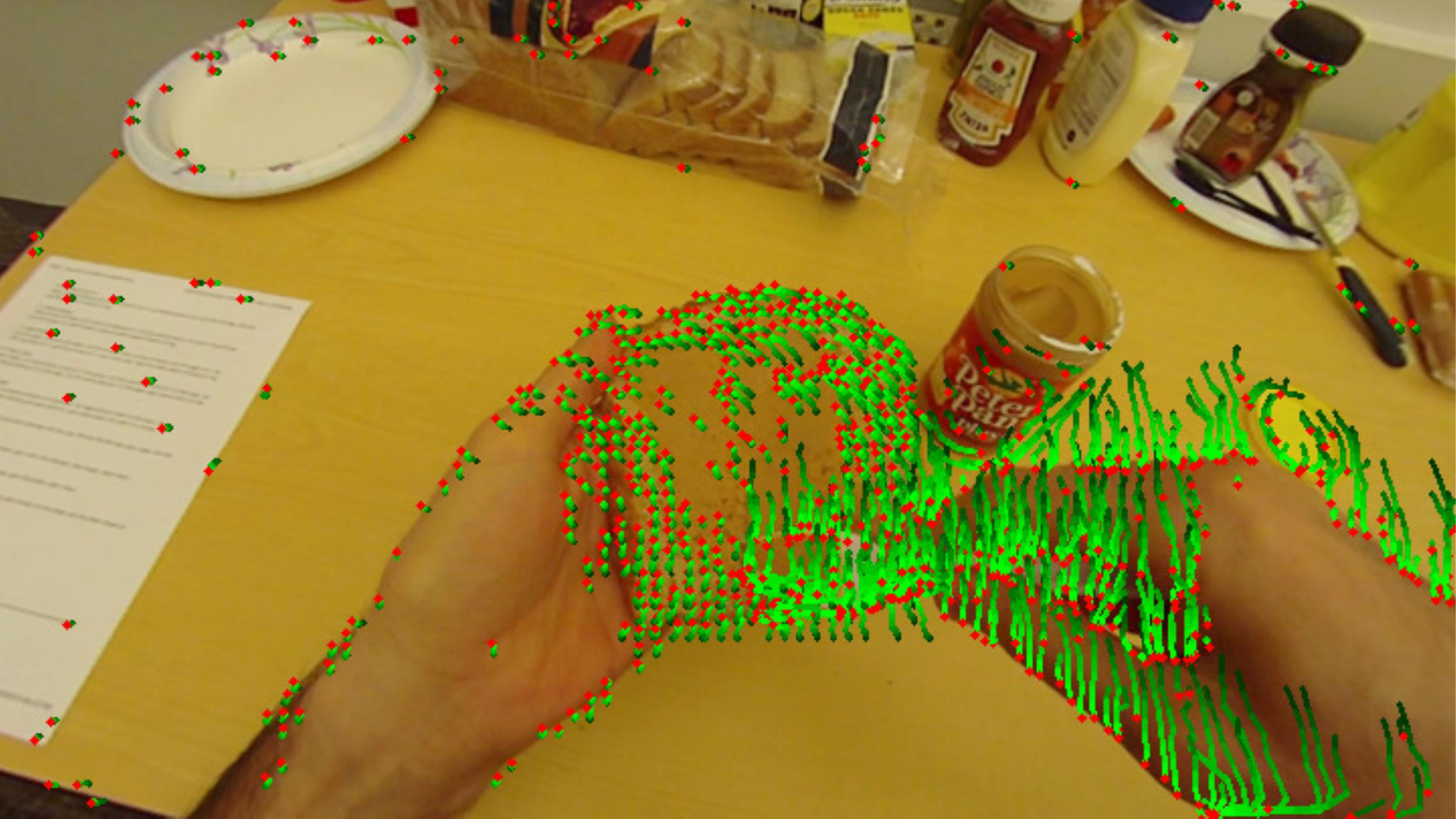}\\ \vspace{0.05cm}
    \caption{The focus of this paper is on recognizing wearer's actions from egocentric videos. Earlier work in this area has suggested complicated image segmentation followed by hand or object recognition (left image). We observe that salient objects (hands or handled objects) in such actions are also the objects moving dominantly with respect to the background and can be captured easily using trajectory aligned features (right image) without any prior image segmentation or hand or object recognition. The example images shown here are from {\sc gtea} database \cite{fathi2011learning}.}
    \label{fig:motivation}
\end{figure}

Our focus in this paper is on recognizing wearer's actions from an egocentric video. Owing to their shakiness, the egocentric videos are significantly more challenging to analyze than third person videos. Action recognition gives structure to such otherwise `wild' videos which can then be used to search, index or browse large number of such videos available on the web today. Action recognition is also usually a first step in many other egocentric applications e.g. video summarization, augmented reality, real time suggestions etc. We follow the popular notation in the field to differentiate between `activity' and `actions'. \emph{Activity} is a high level description of what a person is doing at a particular point of time. The activity is usually composed of many short term \emph{actions}, which are perceptually closer to the gestures performed by the person. For example, while making tea is an activity, picking the jar, opening the lid and taking sugar etc. are the actions. Other types of actions popular in computer vision are sitting, standing, jumping etc.

Egocentric videos are different from their third person counterpart, not only because of the change in camera perspective but also because of change in camera motion profile. Many of the accepted techniques for third person video analysis do not work as is for egocentric videos and the community has been grappling with adapting or developing from scratch, solutions to these fairly standard problems in the new context. Work done in last few years have ranged from simpler problems like object recognition \cite{fathi2011learning, ego_handled_objects, ego_obj2}, activity recognition \cite{fathi2011understanding, fathi2012learning, ryoo2013first, spriggs2009temporal, pirsiavash2012detecting, sundaram2009high, ogaki-cvpr12, ego_ac_recog_cvprw14} to more complex problems like summarization \cite{lee2012discovering, lu2013story, aghazadeh2011novelty}, and social interactions \cite{fathi2012social}. Interesting ideas which exploit special properties of egocentric videos have been proposed for problems like temporal segmentation \cite{poleg2014temporal, kitani2011fast}, frame sampling \cite{grauman-snap-points, egosampling} and hyperlapse \cite{hyperlapse}. Newer areas specific to egocentric vision such as gaze detection \cite{li2013learning} and camera wearer identification \cite{ego_biometrics, egosig} have also been explored.

\begin{figure*}[t]
    \centering
    \includegraphics[width=0.18\linewidth]{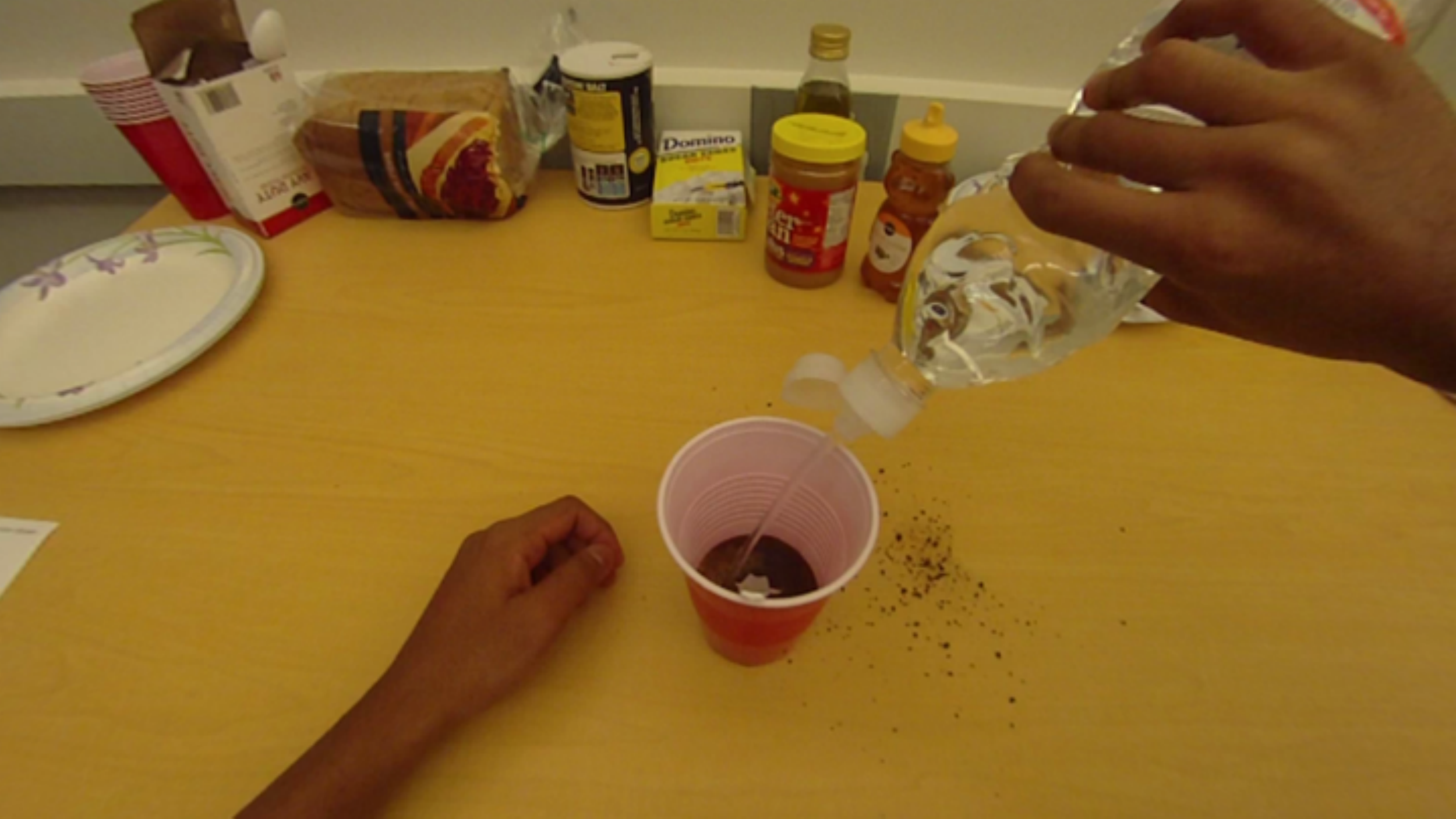}
    \includegraphics[width=0.18\linewidth]{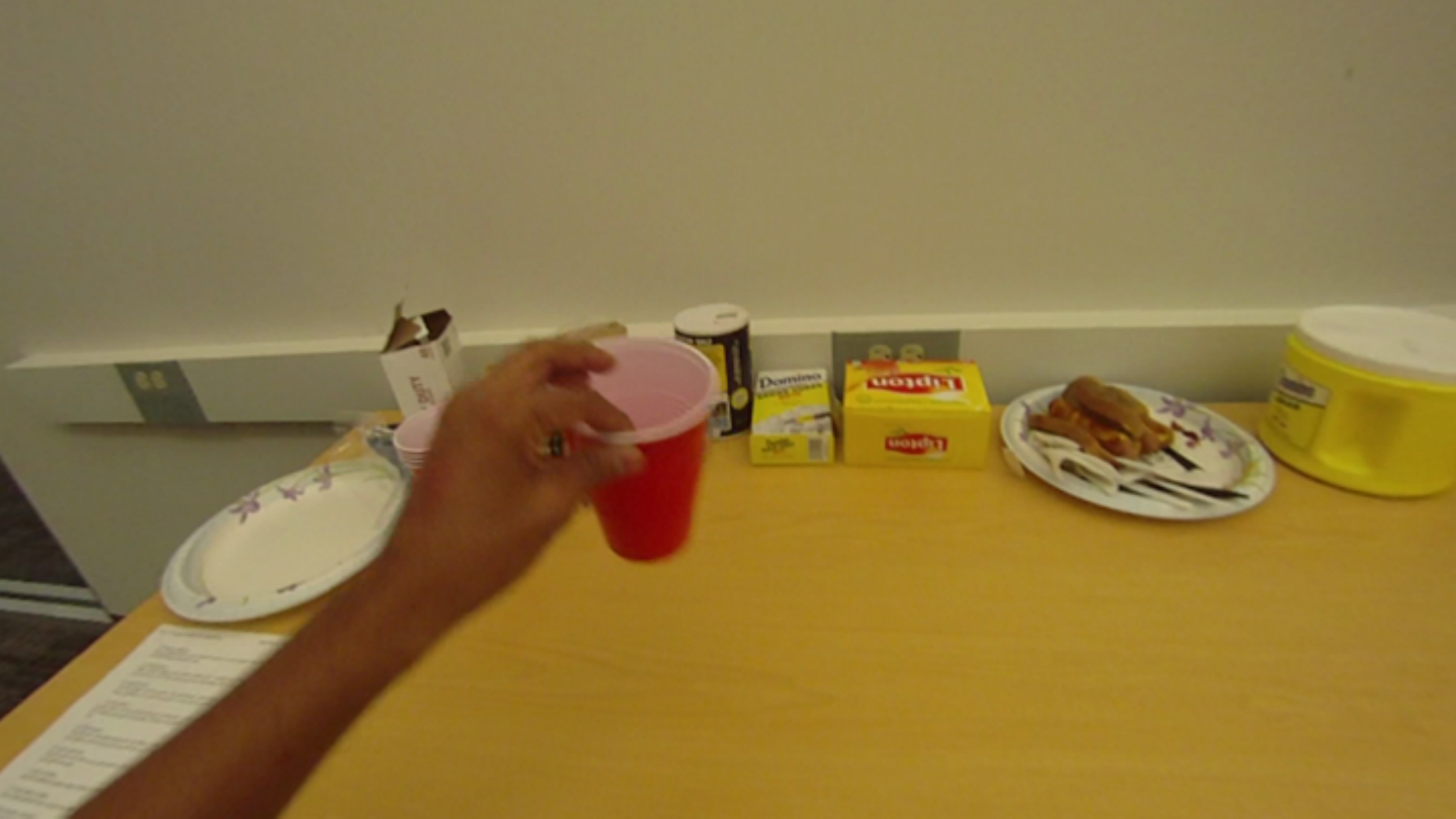}
    \includegraphics[width=0.18\linewidth]{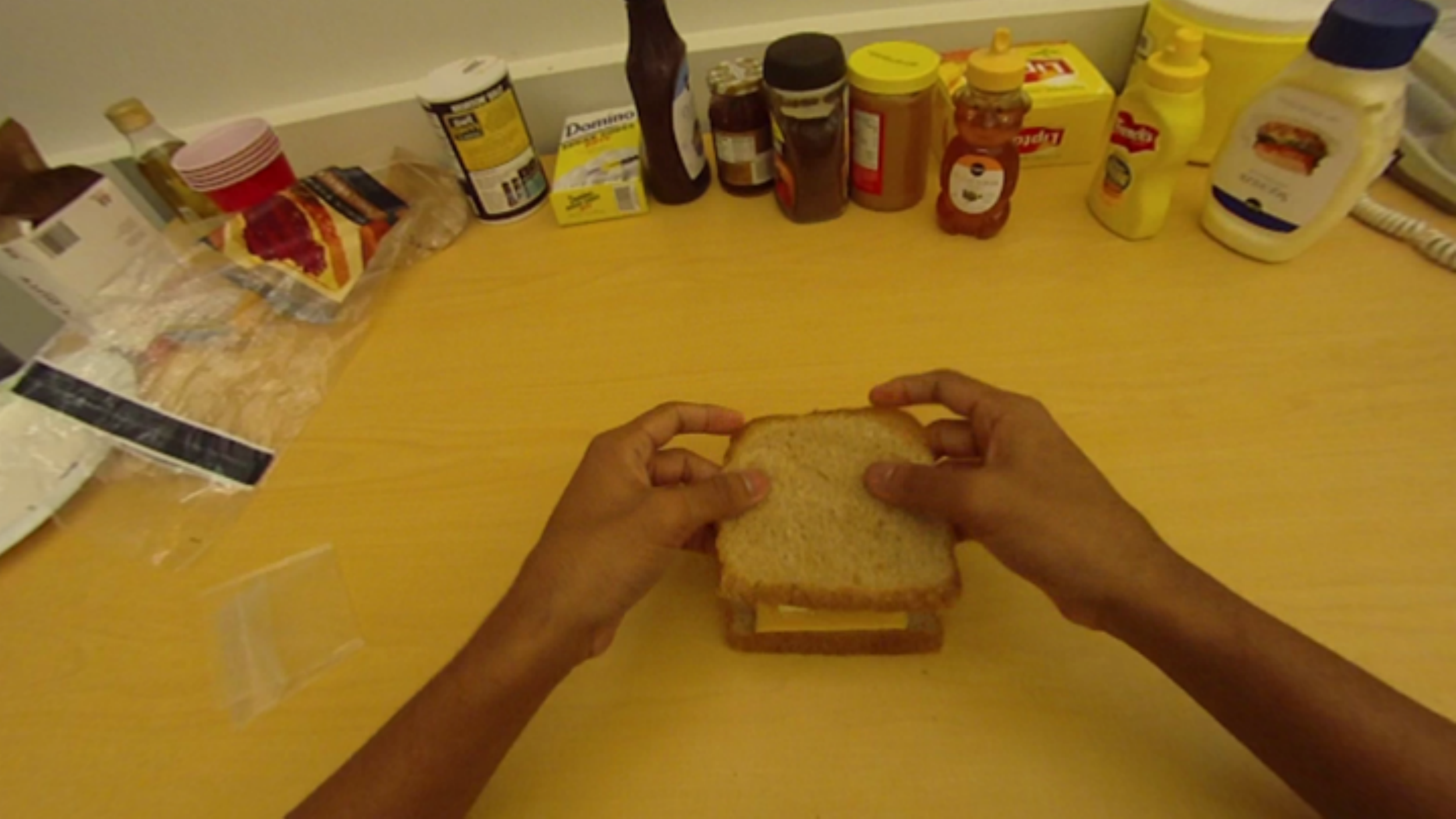}
    \includegraphics[width=0.18\linewidth]{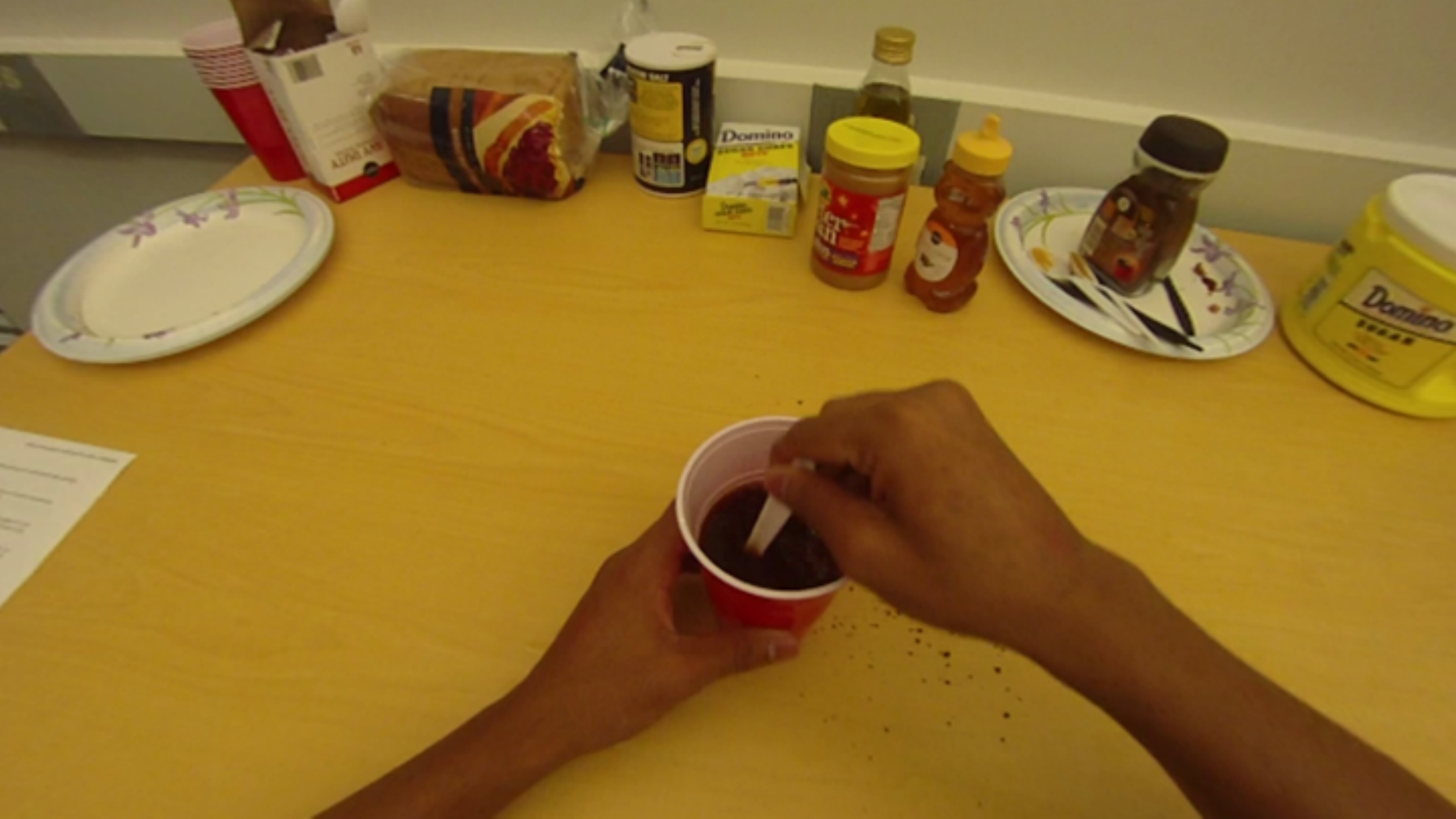}
    \includegraphics[width=0.18\linewidth]{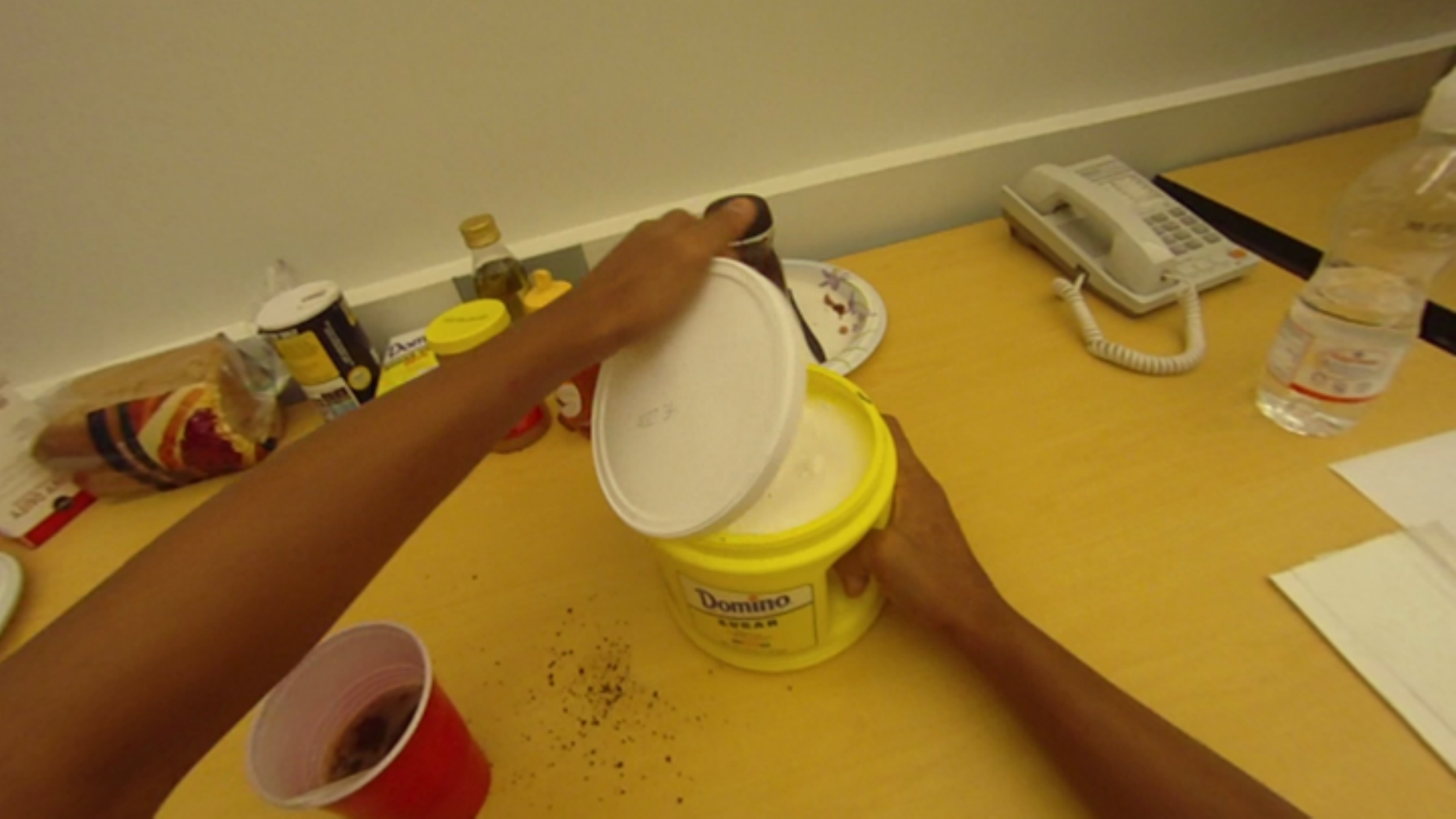}\\[2pt]
    \includegraphics[width=0.18\linewidth]{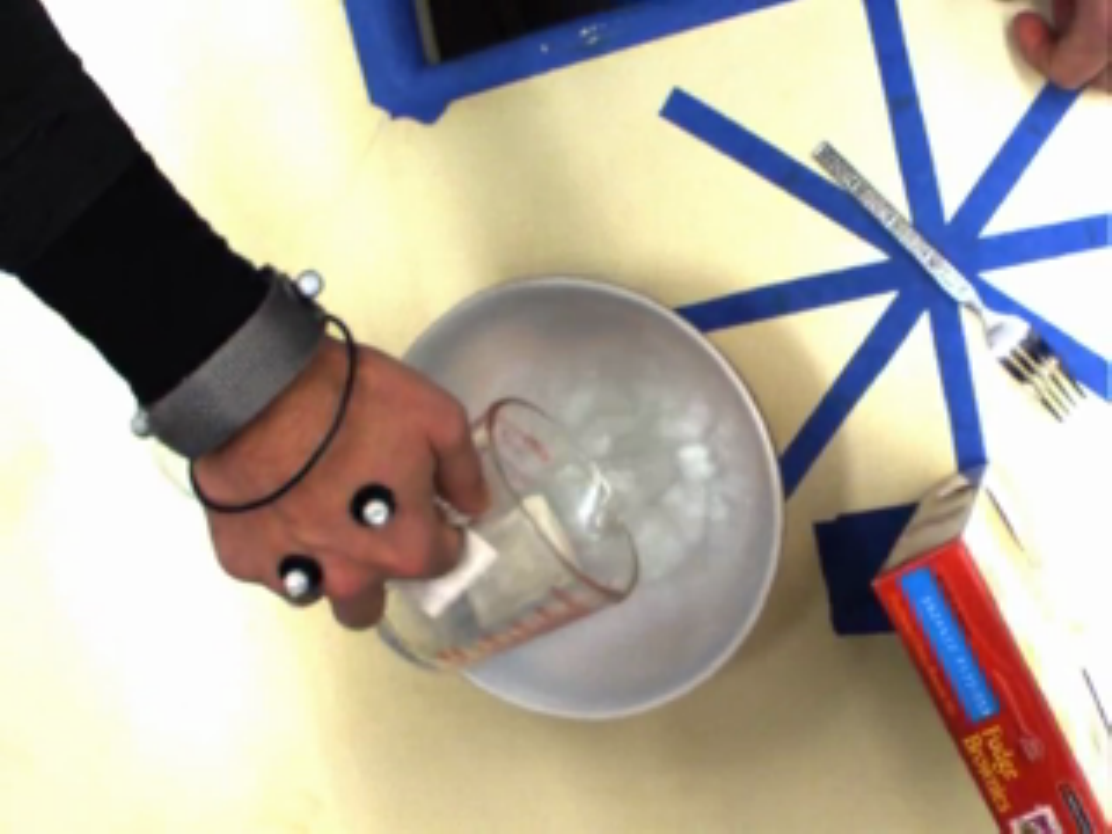}
    \includegraphics[width=0.18\linewidth]{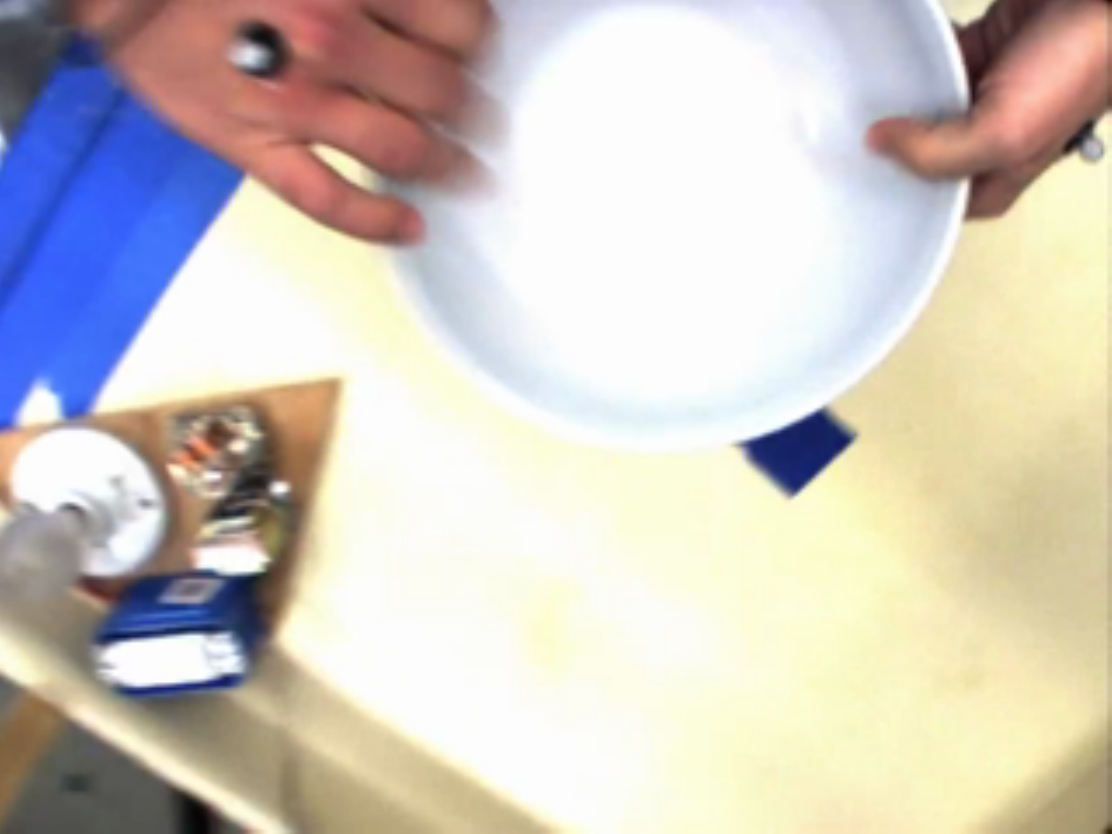}
    \includegraphics[width=0.18\linewidth]{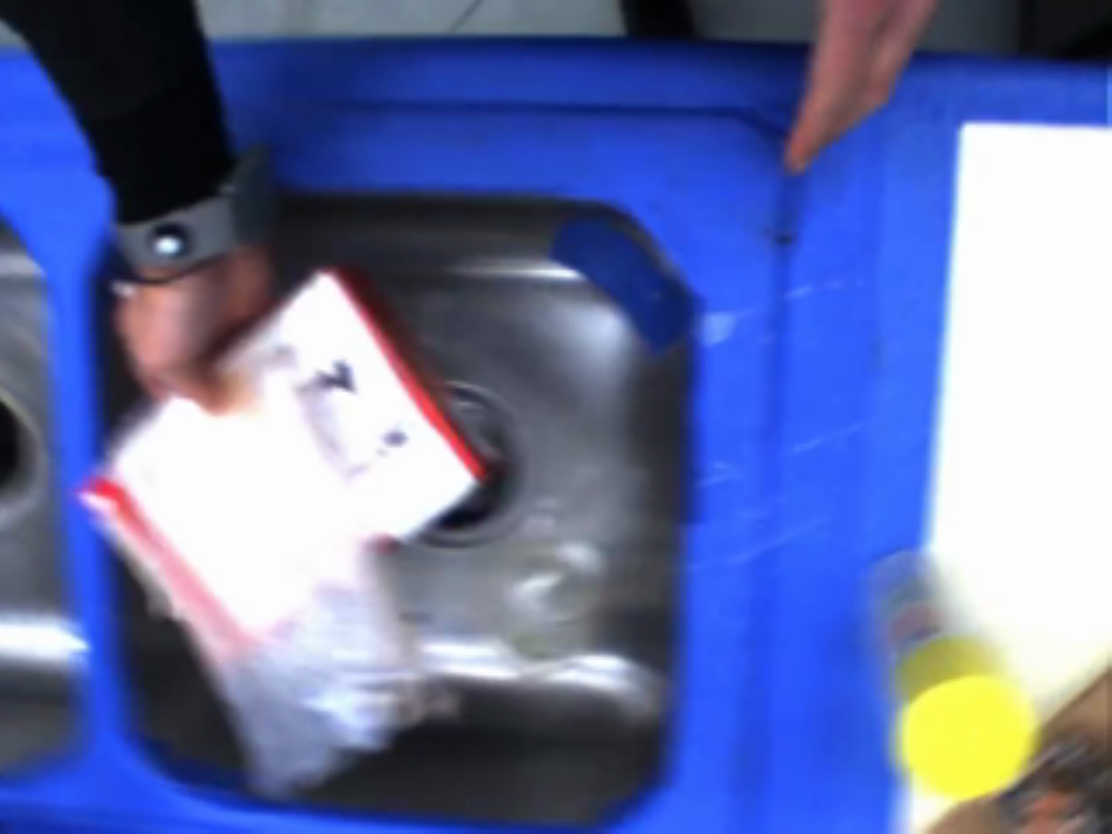}
    \includegraphics[width=0.18\linewidth]{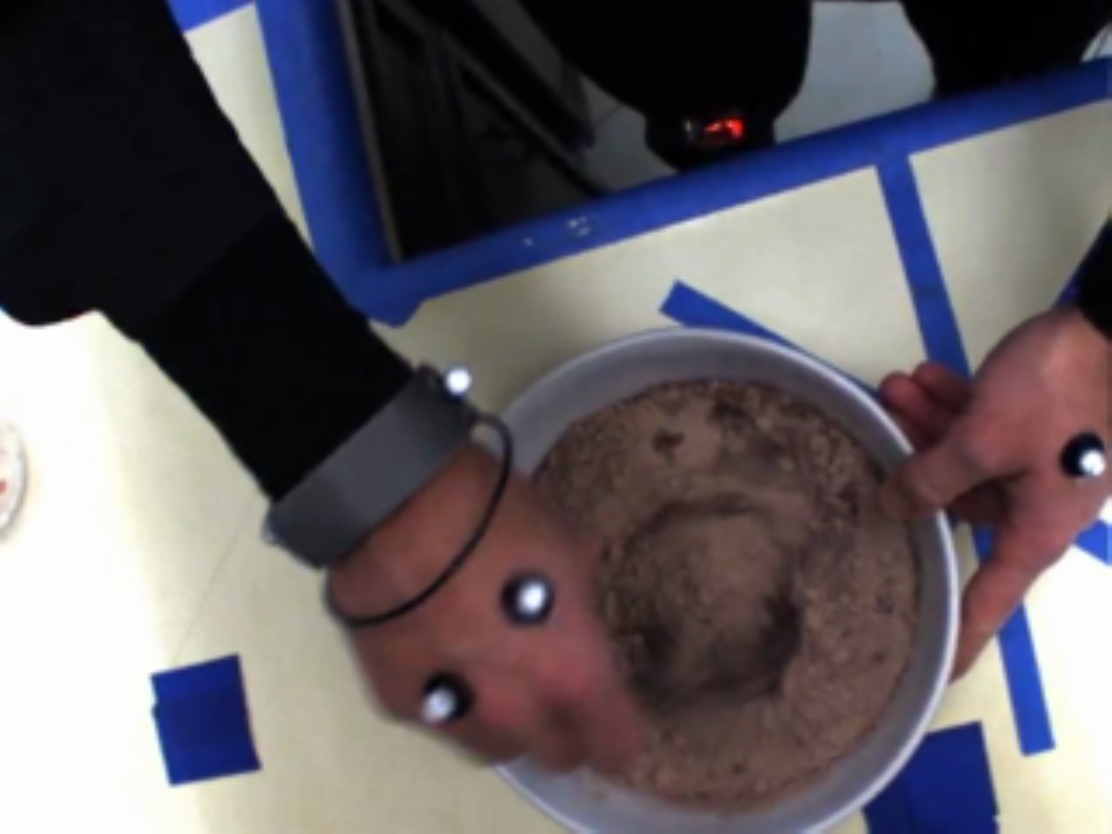}
    \includegraphics[width=0.18\linewidth]{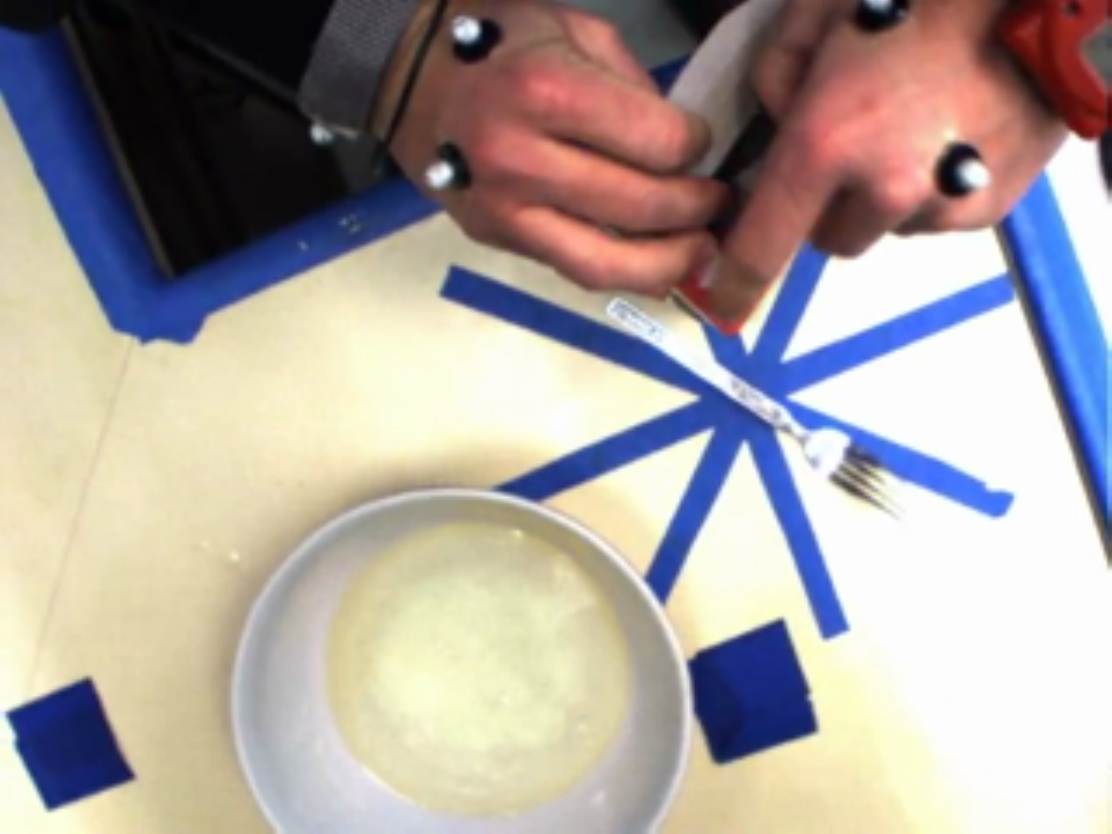}\\[2pt]
    \includegraphics[width=0.18\linewidth]{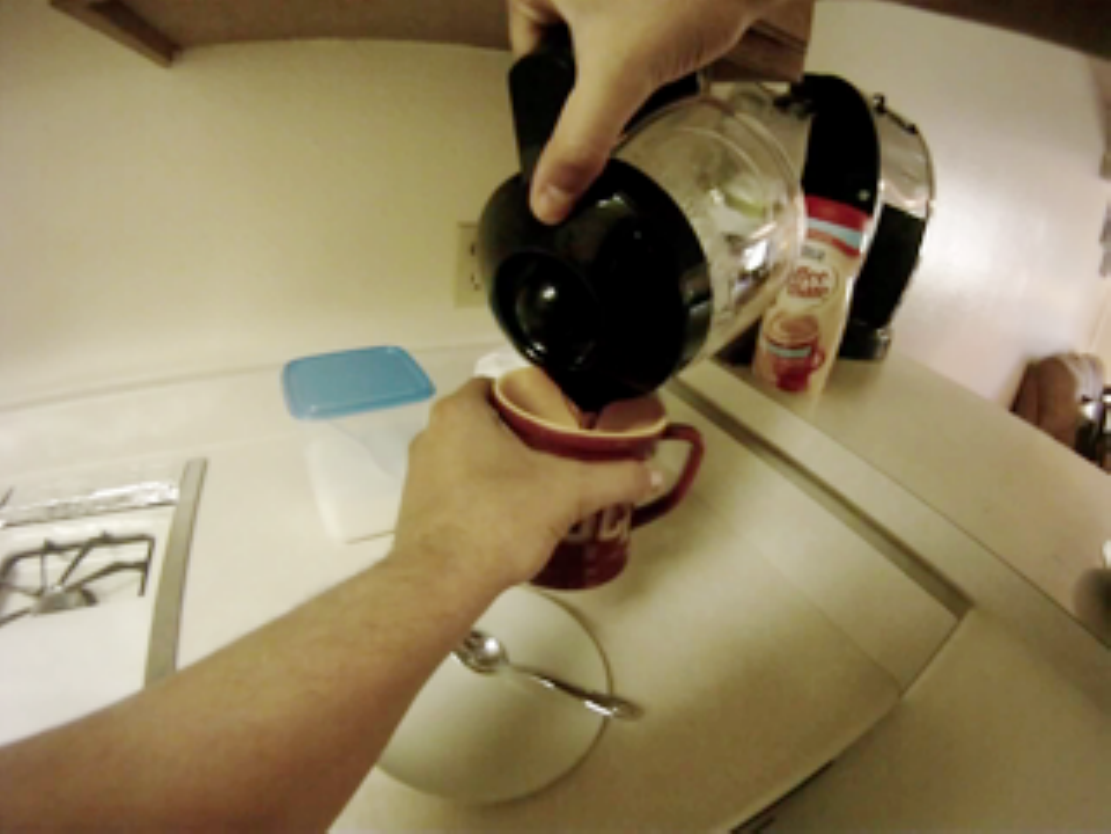}
    \includegraphics[width=0.18\linewidth]{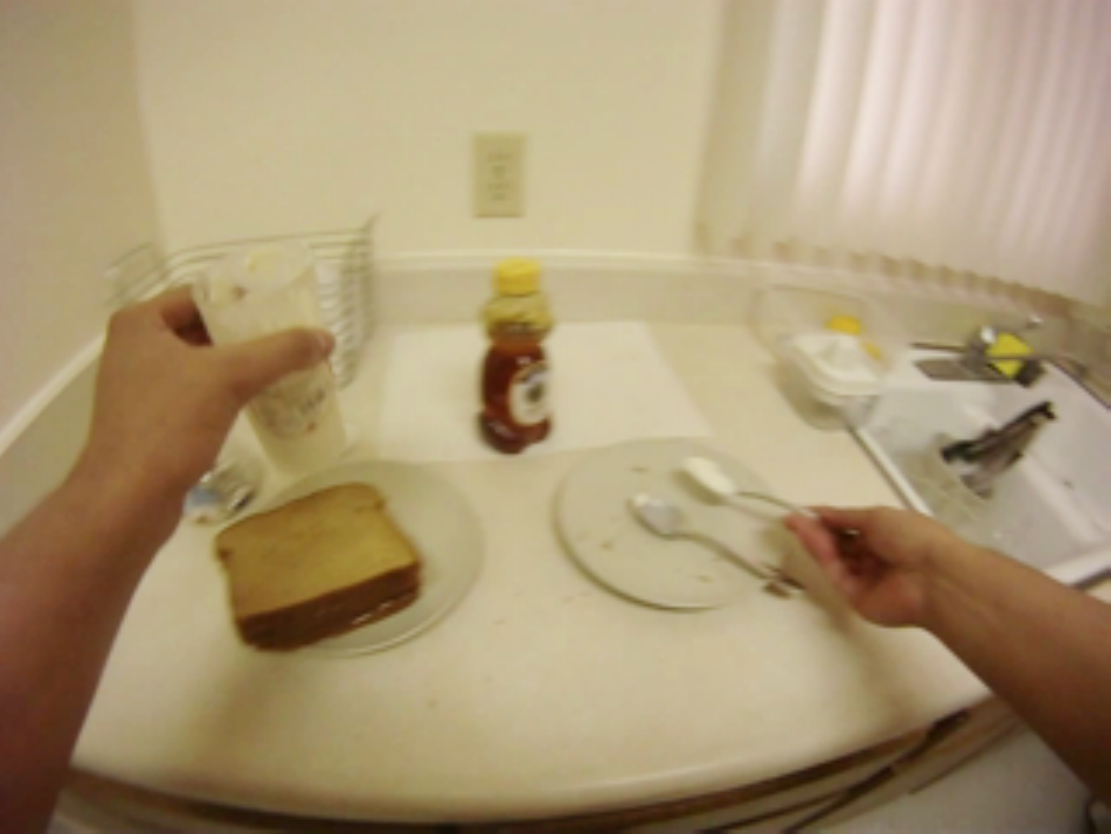}
    \includegraphics[width=0.18\linewidth]{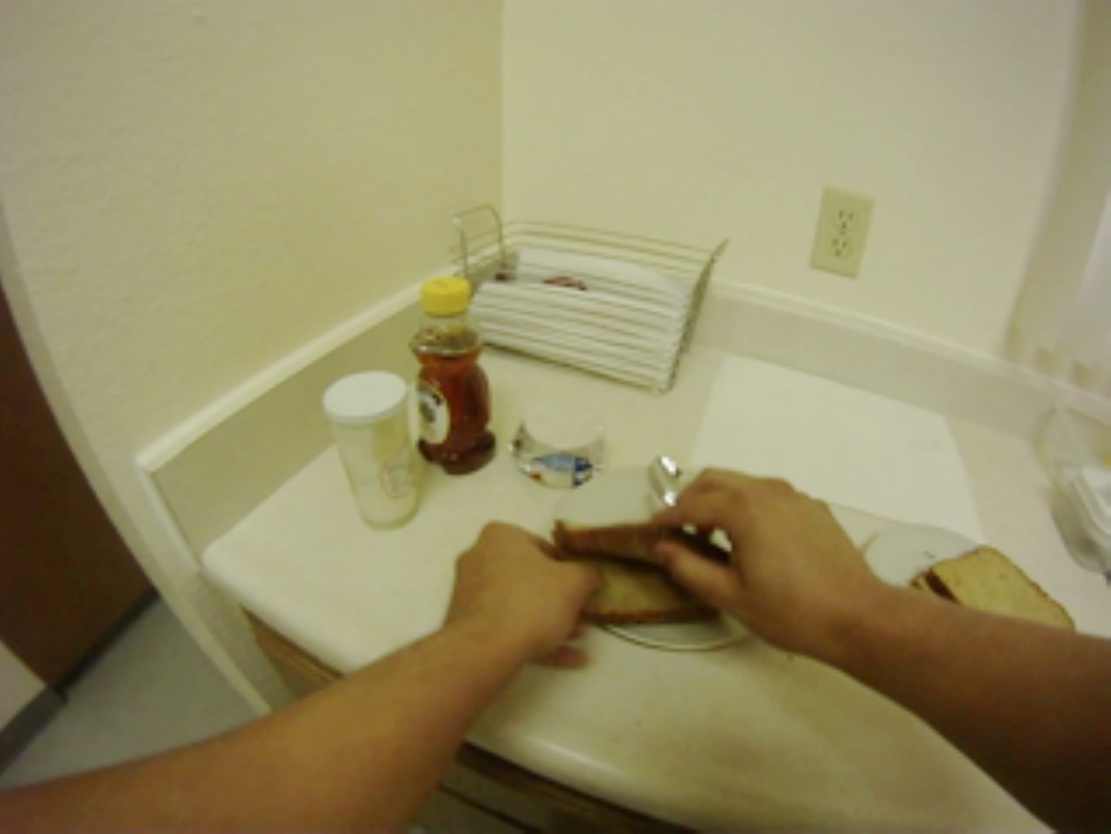}
    \includegraphics[width=0.18\linewidth]{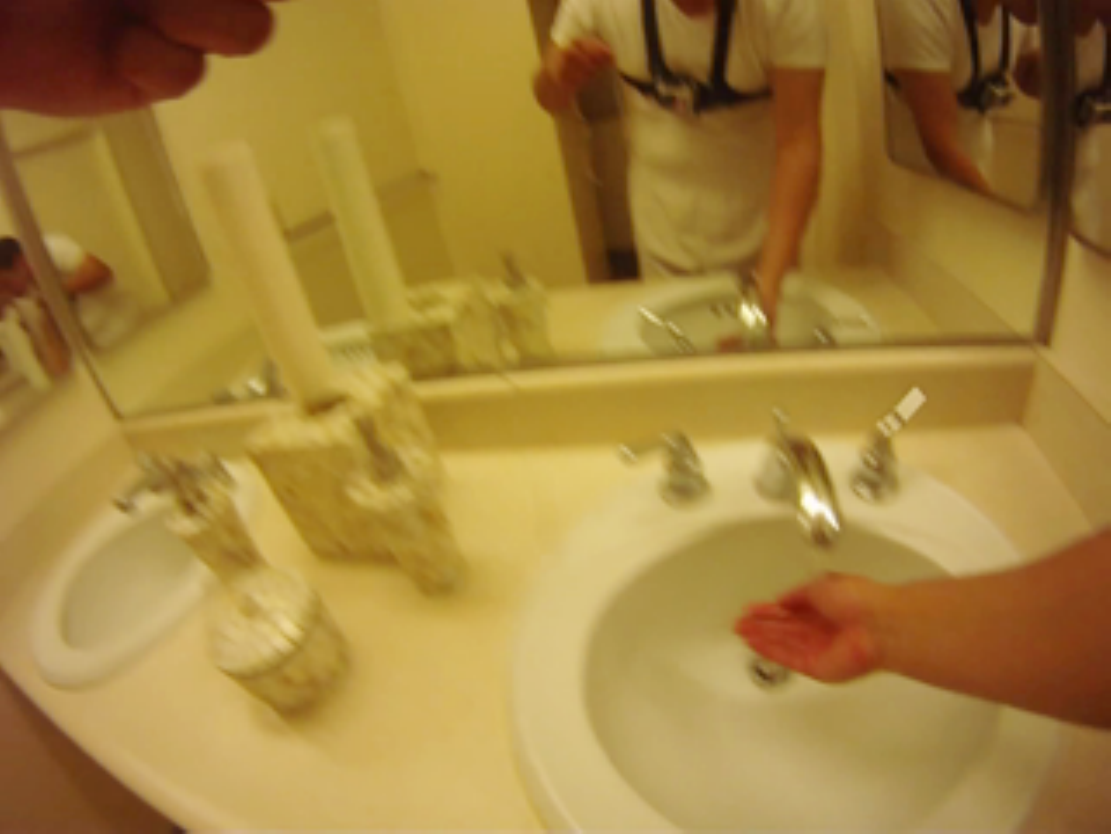}
    \includegraphics[width=0.18\linewidth]{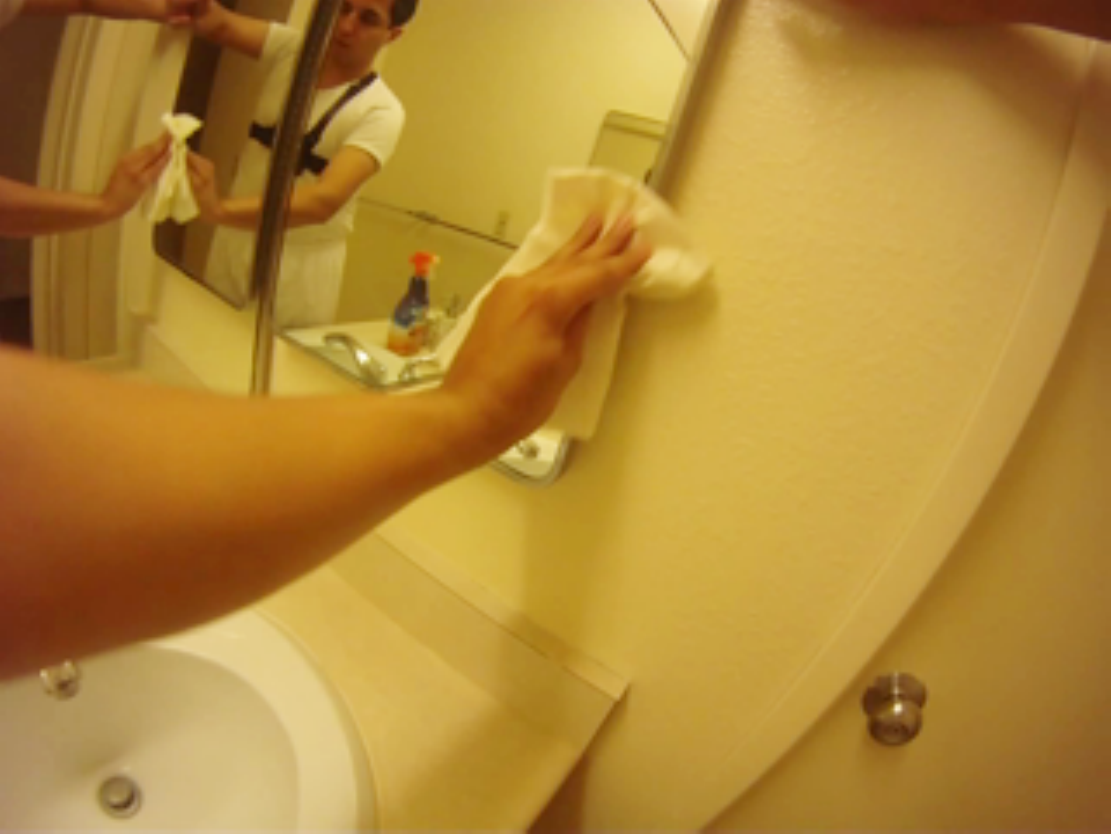}
    \caption{Examples of wearer's action categories we propose to recognize in this paper from different datasets: {\sc gtea}\cite{fathi2011learning} (top row), Kitchen\cite{spriggs2009temporal} (middle row) and {\sc adl}\cite{pirsiavash2012detecting} (bottom row). First, second and third columns across all rows are `pour', `take' and `put' actions respectively. Fourth and fifth columns are `stir' and `open' actions for top and middle rows, and `wash' and `wipe' actions for bottom row. The actions vary widely across datasets in terms of appearance and speed of action. Features and technique we suggest in this paper is able to successfully recognize wearer's actions across different presented scenarios, showing robustness of our method.}
    \label{fig:visual_res}
\end{figure*}

Wearer's action recognition from egocentric video has been a popular problem. The problem is harder compared to regular third person action recognition due to associated unstructured and wild motion of the camera caused by wearer's natural head movement. Different speed of performing actions and widely varying operating environment also causes difficulties. Figure \ref{fig:visual_res}, gives some examples of the actions we are interested in recognizing.

Given the unique perspective of egocentric camera, which makes unavailable, the view of the actor or his/her pose, standard action recognition techniques from third person actions are not applicable as is. Also quickly changing view field in a typical egocentric videos, makes it hard to develop models from foreground or background objects. Therefore, the techniques developed for wearer's action recognition have so far remained independent of work done in third person actions. The earliest work in the wearer's action recognition used global features ({\sc gist}) for the task \cite{spriggs2009temporal}. Later works focussed on objects present in the scene for the recognition \cite{pirsiavash2012detecting, mccandless2013object}. Position and pose of hand is an important cue for action recognition involving object handling and have been explored by the researchers as well \cite{fathi2011learning}. In action categories which does not involve any handled object, researchers have typically exploited the optical flow observed in the video which for an egocentric video is indicative of head motion and is highly correlated with the kind of action being performed by the wearer \cite{kitani2011fast, poleg2014temporal}. Eye-motion and ego-motion have also been used to recognize indoor desktop actions \cite{ogaki-cvpr12}.

Object or hand pose are important cues for wearer's action but detecting them in an egocentric video is a difficult task and the dependence of the action recognition on such explicit detection/recognition effects the overall action recognition accuracy, besides making the system more complex and inefficient. We show in this paper that such prior information is not necessary. We observe that in any egocentric action scenario involving handled object, the dominantly moving objects in the scene are typically hands and handled objects only. Optical flow observed for the background is due to motion of the wearer's head. Such motion causes $3D$ rotation of the camera and can be easily compensated by cancelling frame to frame homography. This leads to a simple algorithm for extraction of hands and objects. We further show that complicated models of hand pose or object recognition are not necessary for the action recognition task and instead simple trajectory based features, combining motion profile and the visual features around these trajectories, alone are sufficient to reach state of the art accuracy. This significantly simplifies the whole processing pipeline for first person action recognition. The simplification also allows to easily generalize the proposed technique to various kinds of actions, not possible with current state of the art, as we show later in the paper. 

\paragraph{Contributions}

We propose a novel representation of egocentric actions based upon simple feature trajectories. Importantly, the proposed features can be computed using tracking alone. The features implicitly capture the visual and motion cues of hands and handled objects and does not require any explicit object detection, hand detection or image segmentation. Use of trajectory aligned features for egocentric action recognition has been proposed for the first time in our paper. We use bag of words to learn the action representation from trajectory based features. Our experiments on publicly available datasets show that the proposed technique improves the state of art by more than $11\%$. We have explored the generalization of our features for action recognition when the wearer's hands or handled objects are not visible. We release an annotated database of $60$ videos for $18$ such action classes performed by different subjects. Interestingly, our technique, not designed for such actions, gives an accuracy of $51.20\%$ on the dataset. Even with significantly simplified compute pipeline, we achieve state of the art results in all the publicly available egocentric datasets. This implies that the proposed features can be used for variety of datasets with significant difference in appearance and actions. We note that none of the earlier state of the art have been shown to apply on ``all the datasets at the same time".

\section{Related Work}

Action recognition has been a popular problem in computer vision. However, this is typically done from a third person view, for example, from a static or a handheld camera. A standard line of work is to encode the actions using keypoints and descriptors. This is done by extending spatial domain descriptors to space-time descriptors. These descriptors are then matched using Euclidean distance or other similar measures.  Some techniques also rely on supervised learning with these descriptor vectors. Some notable contributions in this area includes {\sc stip} \cite{Laptev2005}, {\small 3}{\sc d}-{\sc sift} \cite{Scovanner2007}, {\sc hog}{\small 3}{\sc d} \cite{Klaser2008}, extended {\sc surf} \cite{Willems2008}, and Local Trinary Patterns \cite{Lior2009}. Methods that follow the pipeline of keypoint detection followed by an action descriptor, usually work on a cuboidal video volume. They tend to merge the optical flow and the appearance information from the foreground and objects present in the scene. There have been proposals to demerge these two. Such attempts track feature points in a video and use these trajectories as cues for the action recognition. Some recent methods \cite{Matikainen2009, Messing2009, Sun2009,Sun2010} show promising results for action recognition by leveraging the motion information of trajectories.

Camera motion is very common in real-world videos and poses a significant challenge to any action recognition technique. Wang \etal \cite{wang2011action} propose a descriptor based on motion boundaries, to reduce the interference from camera motion. They compute motion boundaries by a derivative operation on the optical flow field. Thus, motion due to locally translational camera movement is canceled out and relative motion is captured. There have been various improvisation of the technique \cite{mihir_cvpr13,wang2013action,kraft2014accv} decomposing visual motion into dominant and residual motions both for extracting trajectories and computing descriptors.

Egocentric cameras have certain distinct advantages as well as constraints for action recognition. While having much lesser occlusions for the objects in an egocentric video is extremely useful, natural head motion of the wearer brings in additional large camera motion, posing a challenge to any first person action recognition algorithm. Within egocentric community, Spriggs \etal \cite{spriggs2009temporal} proposed to recognize first person actions using a mixture of {\sc gist} \cite{gist} features and {\sc imu} data. Their results confirm the importance of head motion in the first person action recognition. Pirsiavash and Ramanan \cite{pirsiavash2012detecting} attempt to recognize the activity of daily living ({\sc adl}). Their thesis is that the first person action recognition is ``all about the objects", and in particular, ``all about the objects being interacted with". To recognise the objects from first person view, they develop representations including $(1)$ temporal pyramids, which generalize the well-known spatial pyramid to approximate temporal correspondence when scoring a model and $(2)$ composite object models that exploit the fact that objects look different when being interacted with. McCandless and Graumann \cite{mccandless2013object} extend the work by using spatio-temporal pyramid histograms of objects appearing in the action. They devise a boosting approach that automatically selects a small set of useful spatio-temporal pyramid histograms among a randomized pool of candidates. In order to efficiently focus on the candidates, they propose an ``object-centric" scheme that prefers candidates involving objects prominently involved in the actions. \mbox{Fathi \etal} \cite{fathi2011learning} recognize the importance of hands in the first person action recognition. They propose a representation for egocentric actions based on hand-object interactions and include cues such as optical flow, pose, size and location of hands in their feature vector. There is an assumption on the availability of hand, object and background labels in the video. Objects are not always the most important cue in the first person action recognition. In a sports video, when there are no prominent handled object, Kitani \etal \cite{kitani2011fast} use motion based histograms recovered from the optical flow of the scene (background) to recognize the actions of the wearer. Ogaki \etal \cite{ogaki-cvpr12} use eye-motion and ego-motion to recognize indoor desktop actions. Recently, Ryoo \etal have suggested pooled motion features tracking how descriptor values are changing over time and summarizing them to represent an action in the video \cite{ryooPOTfeature}. In a parallel independent work, Li \etal have also proposed a feature descriptor based upon dense trajectories \cite{Li2015cvpr}. However they also use complex patterns like gaze and hand pose which we show are not necessary to reach state of the art accuracy.

\section{Descriptor for First Person Actions} \label{sec:desc}

Motion of handled object and hands is an important cue in first person action recognition. However, unlike previous approaches we believe that segmentation and object recognition is not necessary for first person action recognition. We propose action descriptor based upon feature tracks obtained from egocentric video. The descriptor is an ensemble of different feature vectors obtained from such tracks as well as from visual cues. We construct a bag of words representation separately for each such feature vector. To motivate the importance of each feature vector independently, we explain them sequentially below along with improvement in the the accuracy by adding that feature vector in the descriptor.

\subsection{Baseline: Dense Trajectories}

\begin{figure}[ht]
    \centering
\includegraphics[width=0.48\linewidth]{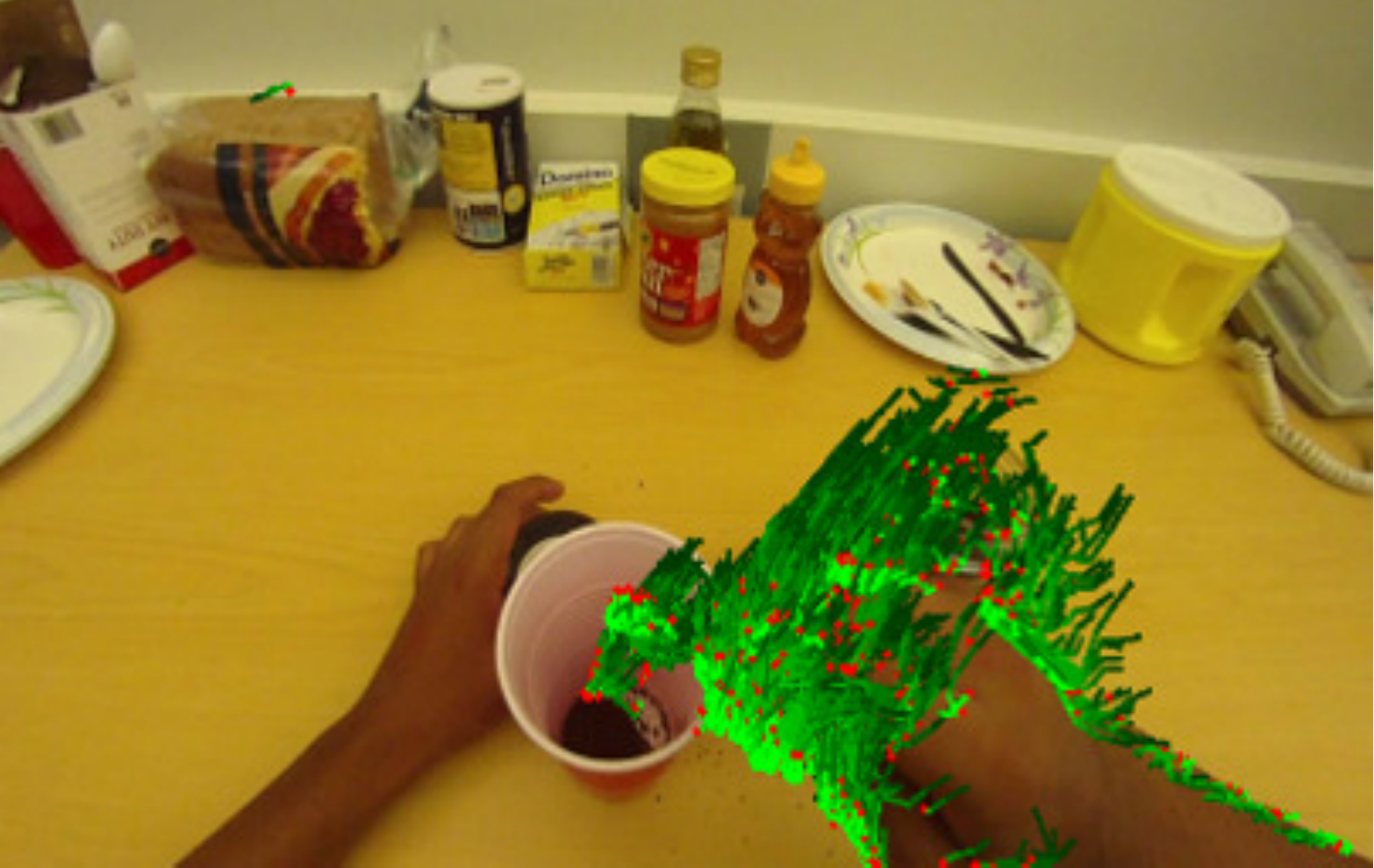}
\includegraphics[width=0.48\linewidth]{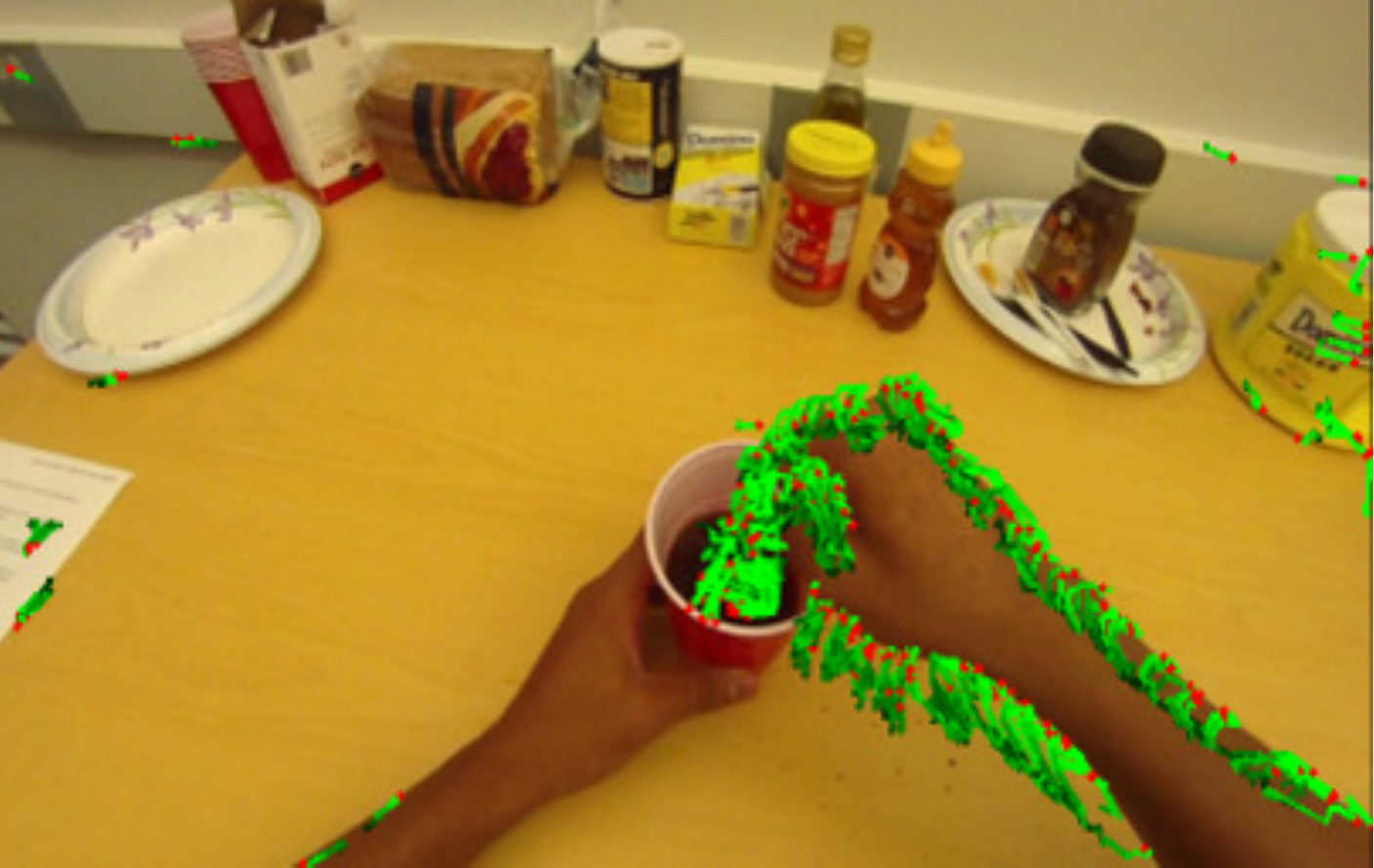}
    \caption{We propose to use the motion cues as well as the visual cues, from the trajectories of object, for first person action recognition. First and second columns show the object and camera trajectories for `pour' and `stir' actions. There is enough information in the cues to classify first person actions. Similar works in egocentric vision use complex image segmentation algorithms to arrive at the labeling of hands and handled objects.}
\end{figure}

In third person action recognition, the constraints of feature representations derived from regularly shaped video volumes is well recognised. Therefore, the newer approaches rely on features computed along the trajectories. Typical keypoint detectors produce sparse feature trajectories effecting the quality of results. Use of feature points sampled on a regular grid as been proposed as a remedial measure. This leads to dense trajectories and improves stability and performance of the algorithms.

We use dense trajectory based feature \cite{wang2011action} as a baseline for our work. As suggested by Wang {\em et al.} \cite{wang2011action}, we extract dense trajectories for multiple spatial scales. Feature points are sampled on a grid spaced by $W$ pixels and tracked in each scale separately. Each point $P_{t} = \left(x_{t}, y_{t}\right)$ at frame $t$ is tracked to the next frame $t+ 1$ by
\[
P_{t+1} = \left(x_{t+1}, y_{t+1}\right) = \left(x_{t}, y_{t}\right) + \left(\mathbb{M} \ast \omega\right)|_{\left(\bar{x}_t, \bar{y}_t\right)}
\]
where $\mathbb{M}$ is the median filtering kernel, $\omega  = \left(u_{t}, v_{t}\right)$ is a dense optical flow field, and $\left(\bar{x}_t, \bar{y}_t\right)$ is the rounded position of $\left(x_{t}, y_{t}\right)$. Tracked points in subsequent frames are concatenated to form a trajectory: $\left(P_{t}, P_{t+1}, P_{t+2}, \ldots \right)$ .

The shape of a trajectory encodes local motion patterns. Given a trajectory of length $L$, we describe its shape by a sequence $S = \left(\Delta P_{t}, \ldots , \Delta P_{t+L-1}\right)$ of displacement vectors $\Delta P_{t} = \left(P_{t+1} - P_{t}\right) = \left(x_{t+1} - x_{t}, y_{t+1} - y_{t}\right)$. The resulting vector is normalized by the sum of the magnitudes of the displacement vectors as
\[
S' = \frac{ \left(\Delta P_{t}, \ldots , \Delta P_{t+L-1}\right)}{\sum_{j=t}^{t+L-1} \lVert \Delta P_j \rVert }.
\]
The vector is referred to as \emph{trajectory descriptor}.

To leverage additional motion and appearance information in dense trajectories, we compute {\sc hog} and {\sc hof} descriptors within a space-time volume around the trajectory. The size of the volume is $N \times N$ pixels and $L$ frames. The volume is subdivided into a spatio-temporal grid of size $n_{\sigma} \times n_{\sigma} \times n_{\tau}$ . We use the default sampling step size of $W = 5$ and $8$ spatial scales spaced by a factor of $1/\sqrt{2}$ and parameters $N = 32$, $n_{\sigma} = 2$, $n_{\tau} = 3$.  Length of a trajectory is set to $L = 15$ frames. For both {\sc hog} and {\sc hof} orientations are quantized into $8$ bins using full orientations, with an additional zero bin for {\sc hof} . Both descriptors are normalized with their $L_2$ norm.

In order to  classify the action at frame $m$, we take a sliding window of size $M+1$ frames and extract dense trajectories within this window. A sliding window centered at frame $m$ consists of $\left( m-M/2,\ldots, m,\ldots, m+M/2 \right)$ frames. In all our experiments, each sliding window consists of $31$ frames ($M=30$). Frames at the border of the video are appropriately padded by reflection. We use Bag of Words ({\sc bow}) model to represent the video segment. Vocabulary for each feature is build separately. For vocabulary construction, we randomly select $10\%$ of training data and then use KMeans clustering and vector quantization for hard vocabulary assignment. \cite{wang2011action} uses vocabulary of size $4000$ for each feature and concatenated histograms are use for classification using a one-\textit{vs}-rest {\sc svm} classifier with $\chi^2$ kernel. In the similar way, we use vocabulary size of $2000$ for all features for all experiments on {\sc gtea} dataset \cite{fathi2011learning}. Later, the histograms corresponding to each feature is concatenated for classification. For classification, we train a one-vs-rest {\sc svm} classifier using $\chi^2$ kernel. The classifier parameters are estimated using $4$ fold cross validation.

The experiments conducted using dense trajectories with {\sc hog} and {\sc hof} descriptors resulted in accuracy of $50.17\%$ and $30.16\%$ respectively, on {\sc gtea} dataset \cite{fathi2011learning}. We give more details in Section \ref{sec:exp}.

\subsection{Motion Cues: Motion Boundary Histogram}

Dalal {\em et al.} \cite{dalal2005histograms} proposed the motion boundary histogram ({\sc mbh}) descriptor for human detection from a moving camera, where derivatives of flow instead of raw optical flow itself is used. In the egocentric videos, the use of the gradient of the optical flow counters the effect of head motion by suppressing the flow of the background. Therefore, we also use {\sc mbh}  in the proposed scheme. For computing the {\sc mbh}  descriptor we compute the spatial derivative of the optical flow field $I_{\omega} = \left(I_{x}, I_{y}\right)$, and orientation information is quantized into histograms, similar to the {\sc hog} descriptor. We then obtain an $8$-bin histogram for each component ({\sc mbh}x and {\sc mbh}y) and normalize them separately with the $L_2$ norm.

The experiments conducted using {\sc mbh}  descriptor resulted in $48.69\%$ accuracy on {\sc gtea} dataset and using {\sc hog}+{\sc hof}+{\sc mbh} improves the accuracy to $50.83\%$. For $L=15$, the length of descriptors are 30, 96, 108 and 192 dimensions for trajectory descriptor, {\sc hog}, {\sc hof} and {\sc mbh} respectively.

\subsection{Action in Reverse: Bi-directional Trajectories}

We observe that human beings can recognize an action even if it is played in reverse. Flow fields, and hence {\sc hof} as well as {\sc mbh} histograms, are different but meaningful when feature is computed in reverse direction. By adding the extra information from reverse playback into the feature allows us to detect the action by using information from both playback directions in one go. We use features from forward and reverse trajectory as if they are obtained from independent trajectories and hence the name `bi-directional' trajectories. The lengths of descriptors obtained from a bi-directional trajectory are same as traditional dense trajectory descriptors described in earlier section. The trajectories obtained from both playback direction are use to build {\sc bow} representation together (instead of separate histogram for each) and hence does not affect the {\sc bow} histogram size.

Using bi-directional trajectories improve the frame level, first person action recognition accuracy from $50.83\%$ to $54.61\%$ on the {\sc gtea} dataset.

\subsection{Handling Wild Motion: Head Motion Cancellation}

\begin{figure}[t]
    \centering
\includegraphics[width=0.48\linewidth]{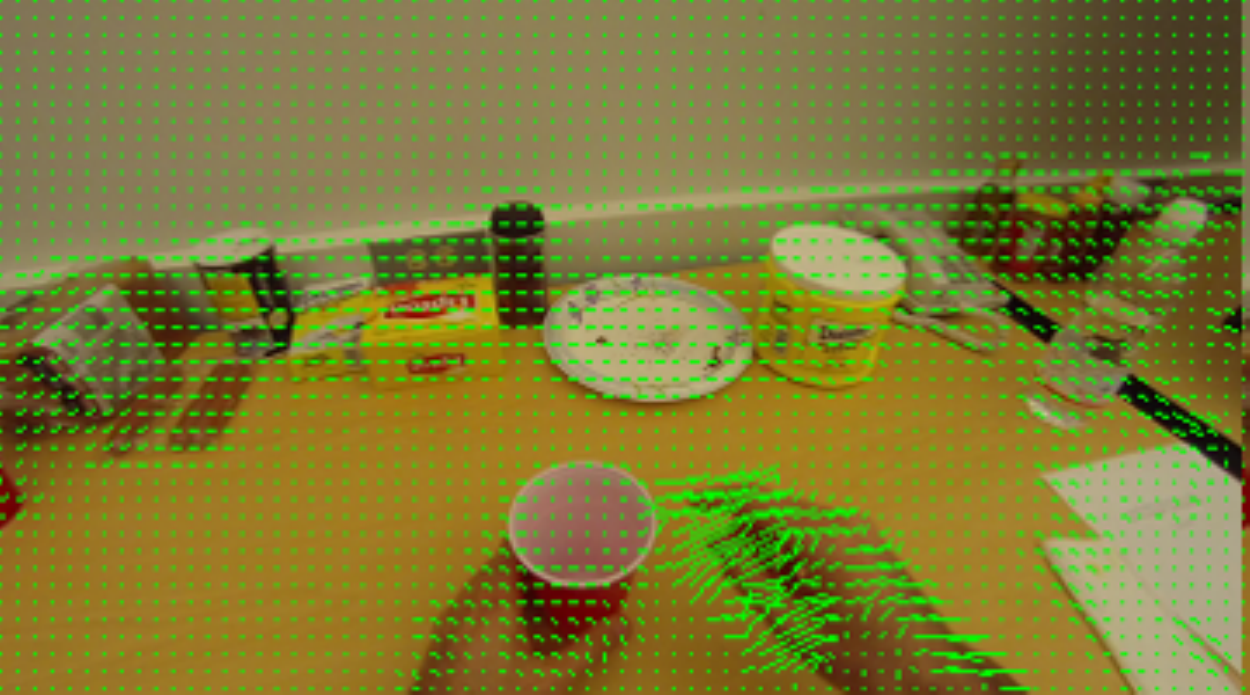}
\includegraphics[width=0.48\linewidth]{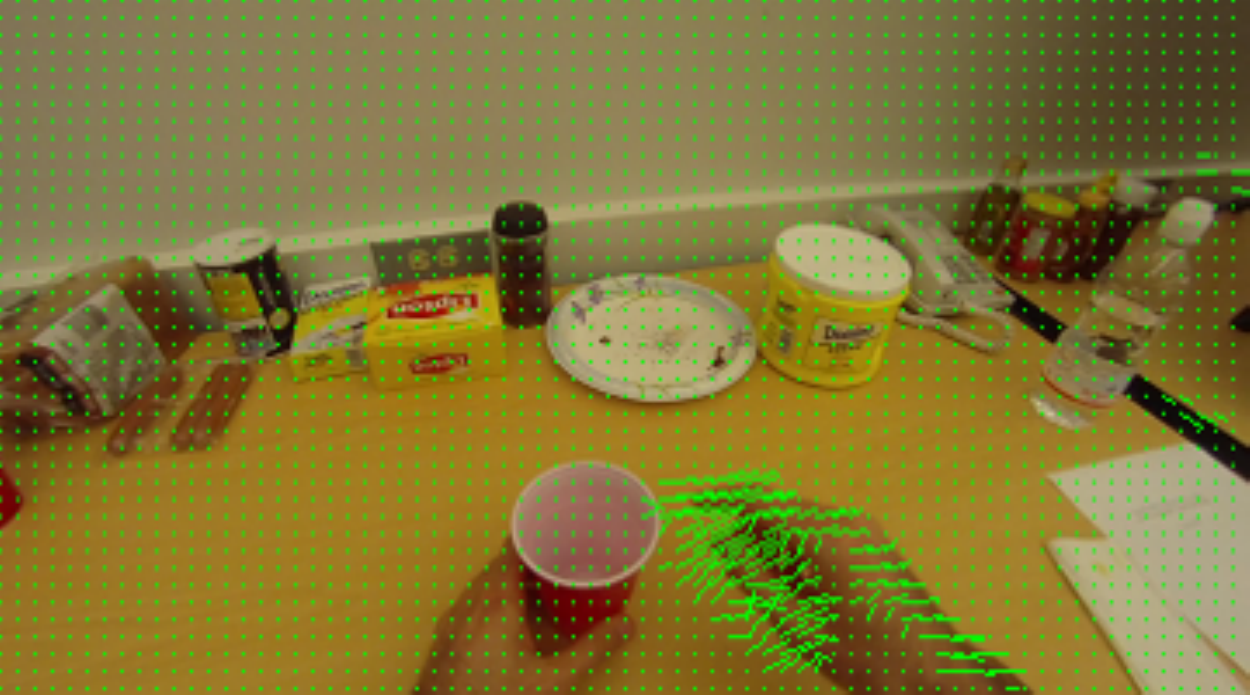}\\

    \caption{Motion of the egocentric camera is due to $3D$ rotation of wearer's head and can be easily compensated by a $2D$ homography transformation of the image. Left: Optical flow overlayed on the frame. Right: Compensated optical flow followed by thresholding. Almost all camera motion has been compensated by this simple technique.}
    \label{fig:dominant_motion_feature}
\end{figure}

The motion of the camera due to head motion of the wearer pollutes the observed trajectories in an egocentric video. By applying head motion cancellation on the flow (see Figure \ref{fig:dominant_motion_feature}), the observed trajectories tend to be smooth and enhance object and hand motion. We model the observed motion due to head movement as $2D$ affine and cancel such motion from trajectory descriptor computed earlier. We observe an improvement in accuracy from $54.61\%$ to $56.87\%$ after cancelling head motion. Interestingly, we observe that camera stabilization as pre-processing also leads to similar gains.

\subsection{Fast and Slow Actions: Temporal Pyramids}

Bag of words representation of trajectory aligned features that we have presented so far ignores the temporal structure of activities. To overcome the limitation we represent features in a temporal pyramid, where top level is a histogram over full temporal extent of the video segment, the next level is the concatenation of two histograms obtained by temporally segmenting the video into two halves (while quantization) and so on. The frame where the trajectory first appears is used to decide the histogram to which it is assigned. All levels of pyramid have the same {\sc bow} histogram size that we have discussed earlier. We obtain a coarse-to-fine representation by concatenating all such histograms together. We use 3 level pyramid for {\sc hog} and {\sc hof} in our experiments which makes feature dimension size  $(2000 \times (1+2+4))$ or $14000$ for {\sc hog} as well as {\sc hof} and $4000$ for {\sc mbh} . Using temporal pyramid further improves the frame level action recognition to $58.50\%$ on {\sc gtea} dataset.

\begin{figure}[t]
    \centering
\includegraphics[width=0.8\linewidth]{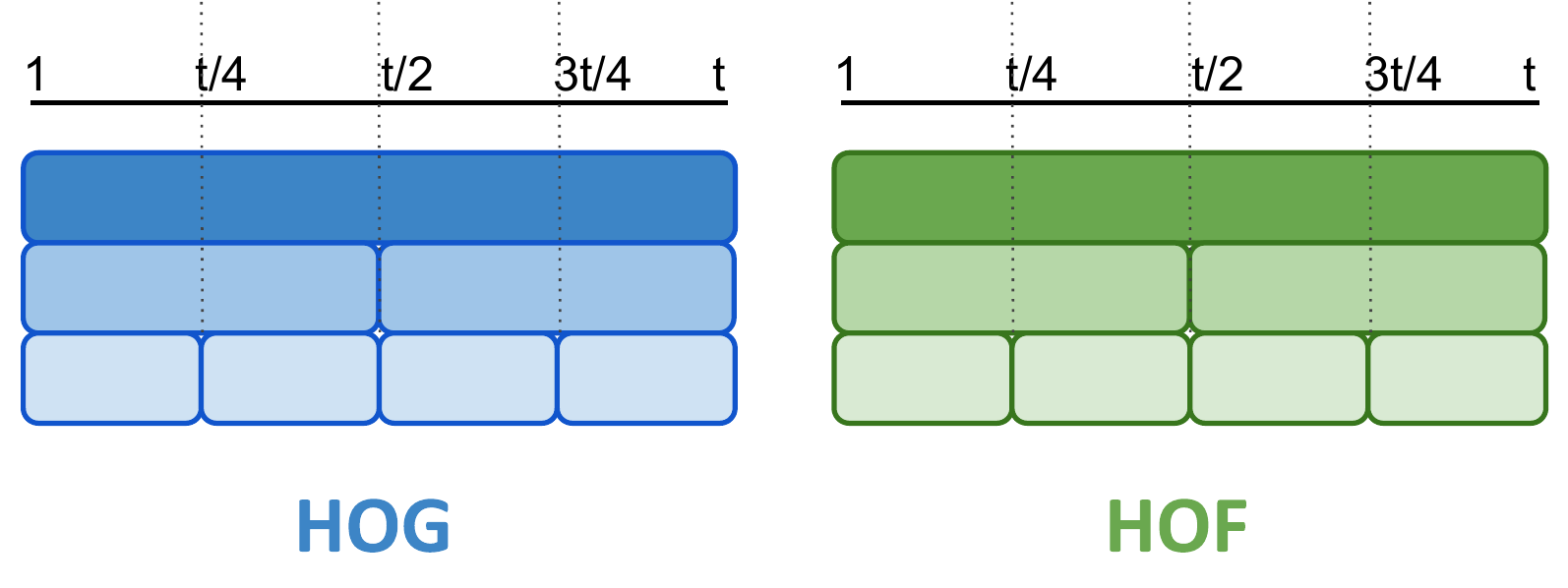}

    \caption{3 level temporal pyramid for a coarse-to-fine BoW representation. Each block is a BoW vector of length $2k$. We use temporal pyramid for {\sc hog} and {\sc hof} features in our experiments.}
    \label{fig:temporal_pyramid}
\end{figure}

\subsection{Kinematic and Statistical features}

As kinematic features, we use local first-order differential scalar quantities computed on the flow field around the trajectories. We consider the divergence, the curl and the hyperbolic terms similar to \cite{mihir_cvpr13}. They encode the physical pattern of the flow which is useful for action recognition. We also use statistics related to trajectories from entire video segment as feature. These features are number of the trajectories, average and standard deviation for $x$ and $y$ coordinates of the trajectory. Trajectory length as well as net displacement of tracked point in horizontal and vertical directions are also added as features. Number of trajectories heading towards each quadrant normalised by total number of trajectories are also appended to it. Kinematic and statistical features improve frame level action recognition on {\sc gtea} to $60.11\%$.

\subsection{Egocentric Cues: Camera Activity}

Camera motion in an egocentric video is due to motion of the wearer's head and is an important cue for action recognition. We represent the camera motion as global frame to frame $2D$ translation, denoted as $\Delta c_{M}=(\Delta x, \Delta y)$. For a video consisting of $M+1$ frames, a camera activity descriptor $C$ is described by a sequence $C = \left(\Delta c_{1},\ldots, \Delta c_{M}\right)$ of displacement vectors. The vector $C$ is normalized by the sum of the magnitudes of the displacement vectors as
\[
C' = \frac{ \left(\Delta c_{1}, . . . , \Delta c_{M-1}\right)}{\sum_{j=1}^{M-1} \lVert \Delta c_j \rVert }.
\]

We concatenate $C'$, total displacement, displacement average and standard deviation and pairs of kinematic features namely (div, curl), (div, shear) and (curl, shear) to represent camera movement and refer to it as \emph{camera activity feature}. Using camera activity feature improves the frame level, first person action recognition accuracy from $60.11\%$ to $61.23\%$ on {\sc gtea} dataset.


\subsection{Semantically Meaningful Temporal Segmentation using Proposed Features}

\begin{figure*}[t]
    \centering
    \includegraphics[width=0.97\linewidth]{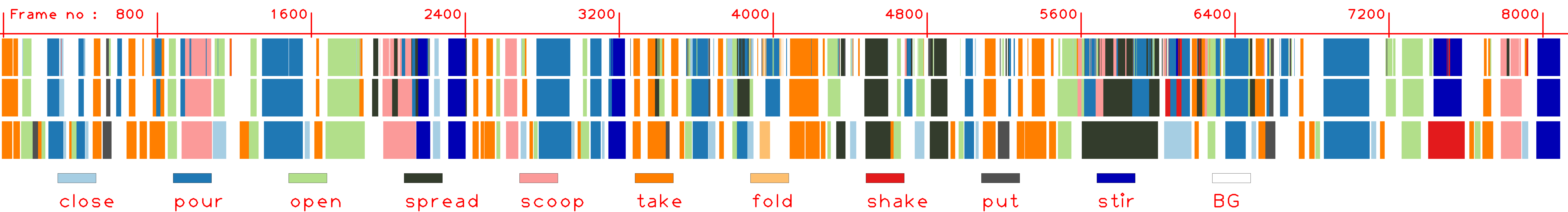}
    \caption{Semantically Meaningful Temporal Segmentation using Proposed Features: Error visualization on all test frames (7 videos) of {\sc gtea} dataset. Each action label has been color coded. We use {\sc mrf} based method for refining predicted label. We assign penalty according to difference in global {\sc hof} histogram of a frame when compared with that of its neighbors. Predicted action labels using classifier score are shown in the top row, action labels after {\sc mrf} based temporal segmentation in the middle row and ground truth action labels in the bottom row.}
    \label{figure:MRF}
\end{figure*}

We have described our features within the context of classification problem so far. However, the features can also be used for temporal segmentation of egocentric videos. For such semantic segmentation we pose our problem as a probabilistic graphical model ({\sc mrf}) where likelihood is derived from classifier score and smoothness prior is used as regularizer. Modelling problem in this way is able to overcome the difficulties in recognizing an action boundary by only likelihood based formulation without prior. We formulate the segmentation problem as follows. Consider a weighted graph where each vertex is a frame which can be labelled with an action label. Each pair of neighboring frames with same action label will be connected by a low weight edge whereas pair of neighboring frames with different action (action boundary) will be connected with a higher weight edge. Neighborhood of each frame is defined as $5$ temporally adjacent frames on both sides (past and future). We assign edge weight using Euclidean distance between global {\sc hof} histograms of $2$ neighboring vertices.
The intuition here is that the change in action between frames should cause a significant change in flow magnitudes and directions in neighboring frames. We proceed to estimate minimum energy cut using $\alpha$-expansion algorithm. We report segmentation accuracy of $62.50\%$ using proposed formulation on {\sc gtea} dataset. Figure \ref{figure:MRF} illustrates the segmentation result and errors using the proposed approach for {\sc gtea} dataset.

\section{Datasets and Evaluation Protocol} \label{sec:dataset}

\renewcommand{\tabcolsep}{0.2cm}

\begin{table*}[t]
\small
    \centering
    \begin{tabular}{lcccccccc}
        \toprule[1.5pt]
        \specialcell{\bf Dataset}  &\specialcell{\bf Subjects} &\specialcell{\bf Videos} & \specialcell{\bf Frames} & \specialcell{\bf Classes}  & \specialcell{\bf Baseline \\ \bf Accuracy \cite{wang2011action}} & \specialcell{\bf State of the art \\  \bf Accuracy} & \specialcell{\bf Our\\ \bf Accuracy} & \specialcell{\bf Temporal \\ \bf  Segmentation} \\ \midrule
        {\sc gtea} \cite{fathi2011learning} & 4 & 28 & 31,253 & 11 & 45.15\% & 47.70\% \cite{fathi2011understanding} &61.23\% & 62.50\% \\
        Kitchen \cite{spriggs2009temporal} & 7 & 7 & 48,117 & 29 & 44.80\% & 48.64\% \cite{spriggs2009temporal} & 59.74\% & 61.42\%\\
        {\sc adl} \cite{pirsiavash2012detecting} & 5 & 5 & 93,293 &  21 & 20.10\% & --- & 31.40\% & 35.16\% \\
        {\sc ute} \cite{lee2012discovering}& 2 & 3 & 208,230 &  21 & 31.78\% & --- & 52.62\% & 55.20\%\\
        Extreme Sports & 60 & 60 & 412,250 &  18 & 43.81\% & --- & 51.20\% & 53.30\% \\
        \bottomrule[1.5pt] \\
    \end{tabular}
    \caption{Statistics of egocentric videos datasets used for experimentation. The baseline accuracy is as achieved using dense trajectory method of Wang {\em et al.} \cite{wang2011action}. The proposed approach uses various trajectory aligned features and improves the baseline as well as the state of the art on all the datasets we tested. The datasets vary widely in appearance, subjects and actions being performed, and the improvement on these datasets validates the generality of suggested descriptor for egocentric action recognition task. Note that originally {\sc adl} dataset has been used for activity recognition and {\sc ute} for video summarization and not for action recognition as in this paper. Therefore, comparative results are not available for these datasets.}
    \label{table:datasets}
\end{table*}

In our work, we use four different publicly available datasets of egocentric videos: {\sc gtea} \cite{fathi2011learning}, Kitchen \cite{spriggs2009temporal}, {\sc adl} \cite{pirsiavash2012detecting} and {\sc ute}  \cite{lee2012discovering}. Out of these, only {\sc gtea} and Kitchen datasets have frame level annotations for the first person actions. For {\sc adl} and {\sc ute} datasets where the similar action level labelling was not available, we selected a subset of the original dataset and manually annotated the short term actions in the parts where a wearer is manipulating some object. Other kind of actions such as walking, watching television etc. are labelled as `background'.  Statistics related to datasets are shown in Table~\ref{table:datasets}.

\paragraph*{GTEA dataset}

This dataset consists of $28$ videos, captured using head mounted cameras. There are $4$ subjects, each performing $7$ long term activities in a kitchen. Each activity is approximately $1$ minute long. We follow experimental setup of  \cite{fathi2011understanding} and use videos of subject `S2' for testing and others for training. There are $11$ action classes viz., `close', `pour', `open', `spread', `scoop', `take', `fold', `shake', `put', `stir', and `background'.

\paragraph*{Kitchen dataset}

The original dataset consists of videos of $43$ subjects performing $3$ activities, captured using head mounted camera and {\sc imu}s. Camera point of view is from top, and severe camera motion is quite common. Similar to  \cite{spriggs2009temporal}, we select 7 subjects from `Brownie' activity, train using videos of 6 subjects and test on the video of remaining subject. There are $29$ classes of actions in this dataset. We selected $7$ subjects ($12$ videos) and manually annotate the short term actions. Each activity is of length in the range of $3$ to $7$ minutes. There are $14$ action classes viz., `close', `pour', `open', `spread', `scoop', `take', `grate', `put', `stir', `crack', `spray', `unwrap', `cut', and `background'.

\paragraph*{ADL videos dataset}

The original dataset consists of videos of $20$ subjects performing $18$ daily life activities, captured using chest mounted camera with $170$ degrees of viewing angle. We selected $5$ subjects and manually annotated the short term actions with $21$ action labels. Similar to  \cite{fathi2011understanding}, we use videos of one subject for testing and the rest for training.
The action classes are `stir', `cut', `shake', `swtichon/off', `take', `open', `close', `fold', `put', `flip', `pour', `wash', `write', `scoop', `wipe', `wear', `tear', `dip', `spray', `type' and `background'.

\paragraph*{UTE dataset}

Original {\sc ute} dataset \cite{lee2012discovering} contains $4$ videos captured from head-mounted cameras.  Each video is about $3$ to $5$ hours long, captured in a natural, uncontrolled setting. We select $3$ parts where hand motion is dominant from $2$ subjects 
 and manually annotate the short term actions. The action labels are `stir', `cut', `shake', `swtichon/off', `take', `open', `close', `fold', `put', `flip', `pour', `wash', `wipe', `tear', `tap', `mix', `peel', `scurb', `rub', `move' and 'background'.

\paragraph*{Extreme Sports}

\begin{figure}[t]
    \centering
    \includegraphics[width=0.4\linewidth]{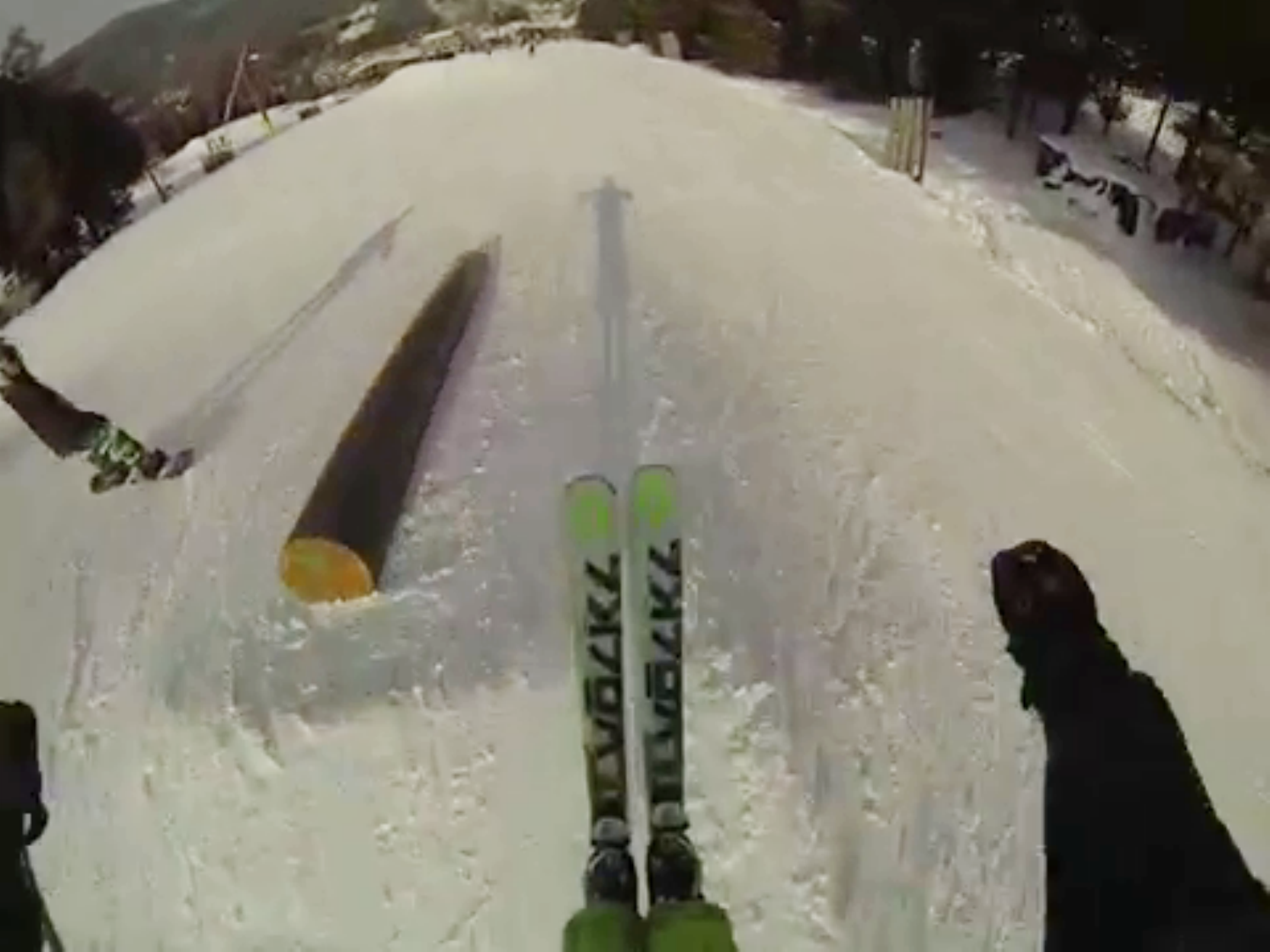} \;
    \includegraphics[width=0.4\linewidth]{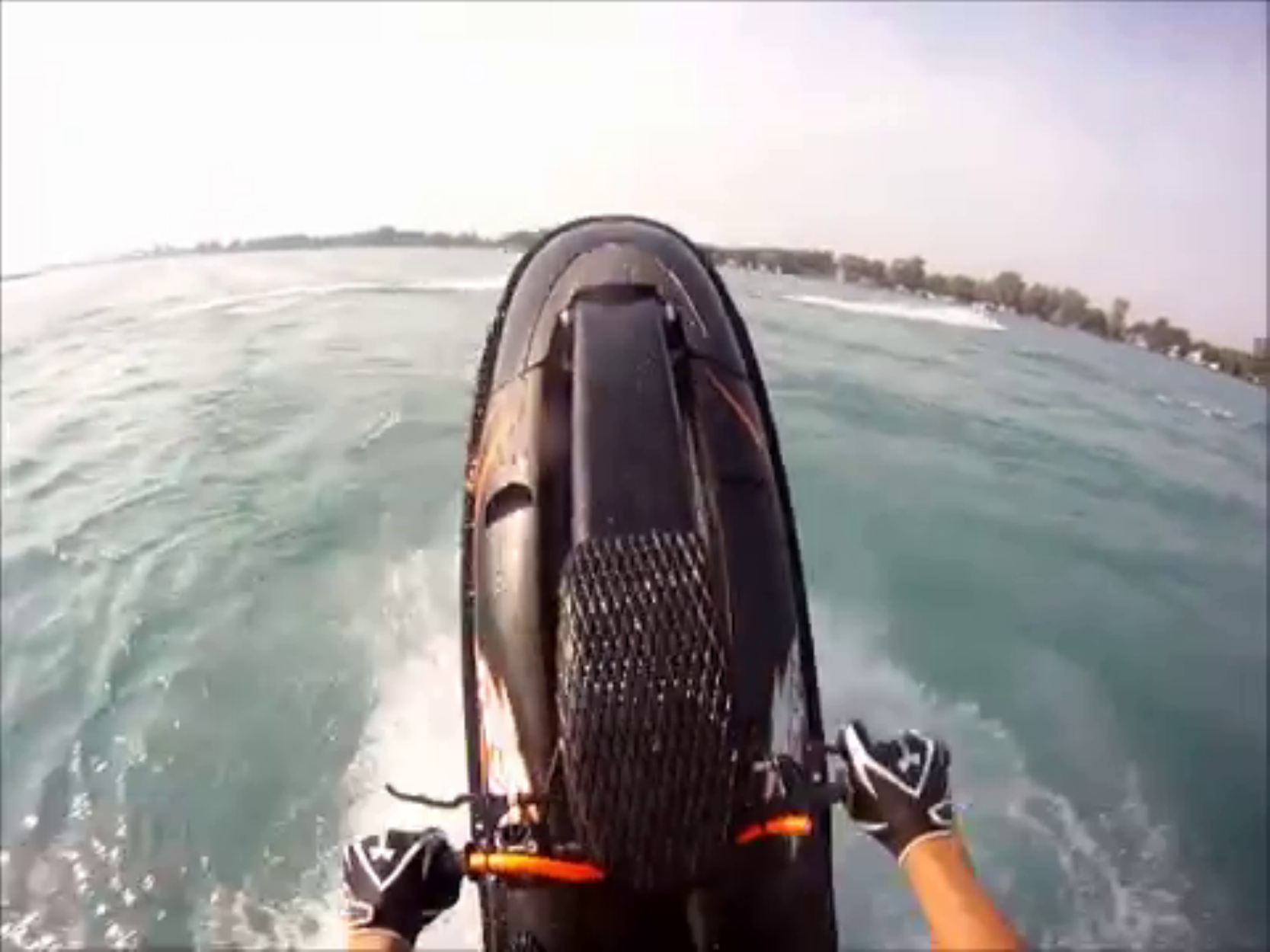} \\ \vskip 0.75em
    \includegraphics[width=0.4\linewidth]{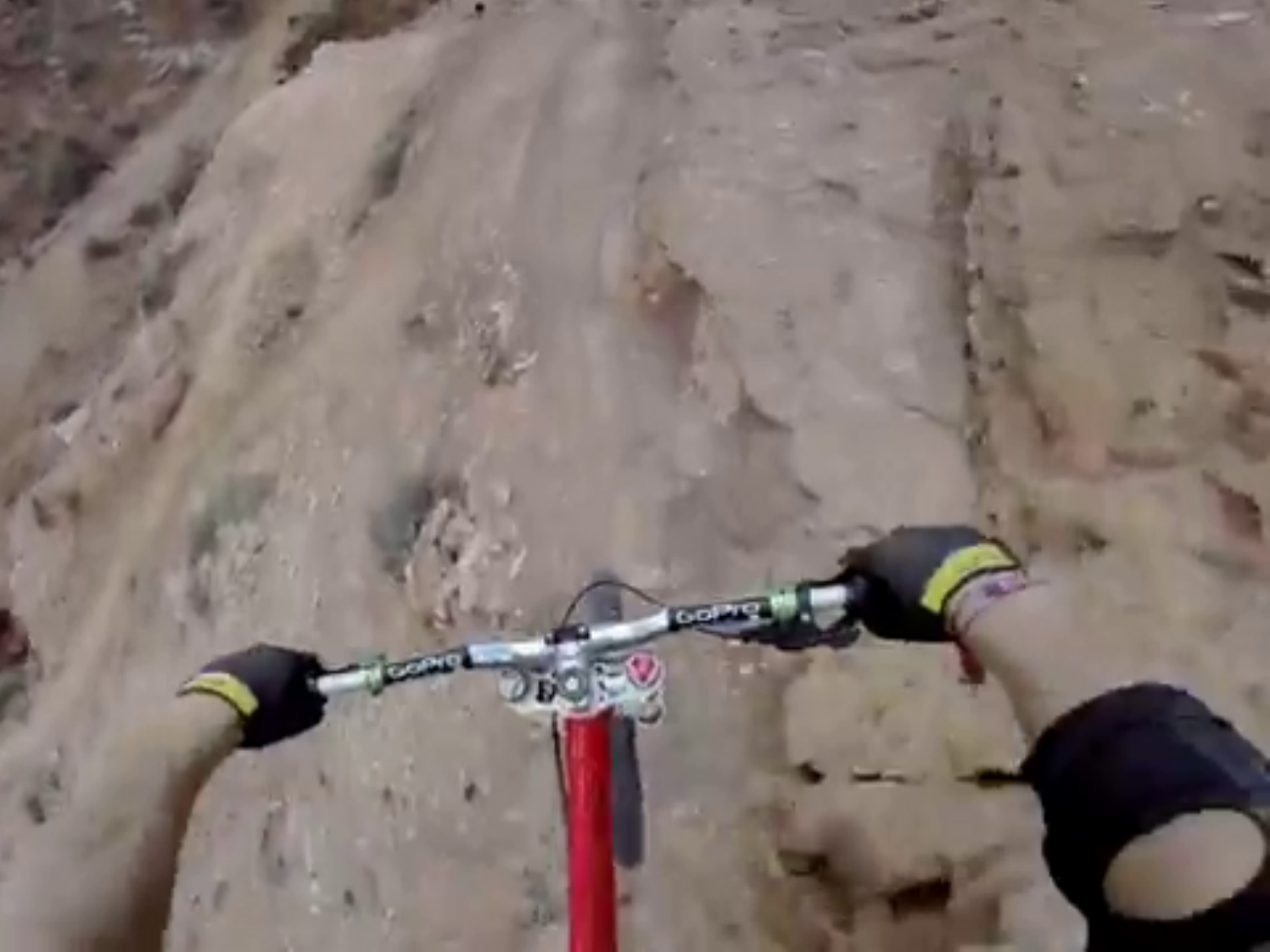} \;
    \includegraphics[width=0.4\linewidth]{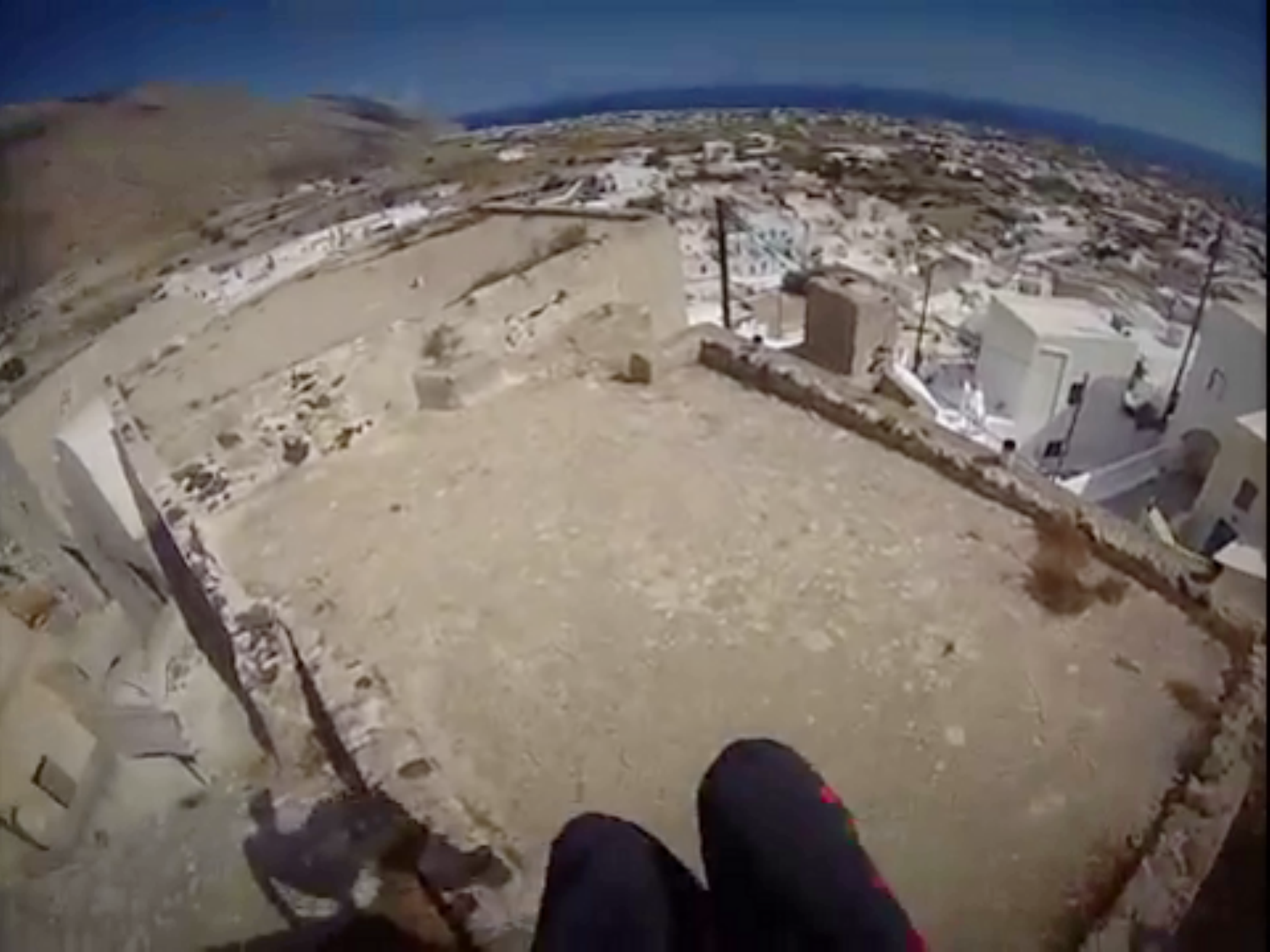}
    \caption{Sample frames from the `Extreme Sports' dataset introduced by us. The figure shows examples for `jump' action in different sports categories: ski, jetski, mountain biking and parkour. Note the variations among the samples which makes the dataset extremely challenging for action recognition task.}
    \label{fig:extreme}
\end{figure}

Most of the egocentric action database we have come across contains actions where wearer's hands or objects are visible. We are also interested in the performance of our features when such cues are not available. Kitani {\em et al.} \cite{kitani2011fast} have suggested unsupervised clustering of such actions for sports videos but the dataset provided by them is quite small (6 categories each with only 1 video). We are introducing a new bigger dataset of similar actions with this paper. We refer to the dataset as `Extreme Sports'. The dataset contains, $60$ videos, amounting to nearly $8$ hours, from $5$ extreme sports categories (mountain biking, jetski, skiing, speedflying and parkour). We have annotated the videos manually into $18$ short term ego-actions similar to Kitani {\em et al.} \cite{kitani2011fast}: `forward', `bumpy forward', `curve-left', `curve-right', `turn-left', `turn-right', `left-right', `jump', `slide-stop', `run', `walk', `roll', `flip', `climb', `vault', `lift', `fly' and `spin'. Figure \ref{fig:extreme} shows some samples from the database.

These sports categories are very different from each other in terms of terrain, nature and types of actions. Due to fast movement nature of extreme sports severe camera shake and motion blur are very common.
We selected first person action classes similar to ego-actions used in  \cite{kitani2011fast}. Each video is captured using head-mounted cameras in diverse terrain (mountain, snow, river, sea, air), weather and lighting. Due to fast movement nature of extreme sports severe camera shake and motion blur are very common.


\subsection*{Evaluation Protocol}

We consider short term actions performed by different subjects while performing different activities. Speed and nature of actions vary across subjects and activities (e.g., consider the action `open' in two scenarios, `open' water bottle and `open' cheese packet).
Formally, classification accuracy for first person action recognition task is defined as number of frames (or video segment) classified correctly divided by total number of frames (or number of video segments) in the videos used for testing. Frame level action recognition is important for continuous video understanding. This is also crucial for many other applications (e.g., step-by-step guidance based on wearer's current actions). We also evaluate our method for action recognition at video segment level. In this case, there is only one action in each video segment. However, length of the segment is not fixed. In this setting, we have an approximate knowledge of action boundaries which naturally improves action recognition results. Segment level action recognition is different from temporal segmentation as each segment is independent of each other. For temporal segmentation we perform labelling of each frame without explicit knowledge about action boundaries.
\section{Experiments and Results} \label{sec:exp}



We first present our experiments and analysis of the proposed action descriptor on {\sc gtea} dataset to bring out salient aspects of the suggested approach. Experiments with other datasets have been described later. Note that these datasets are quite different from each other, and performance improvement on all these datasets compared to the current state of the art show the generality of our features.

Duration of action in all these datasets vary from few frames to few hundred frames. Size of sliding window plays a crucial role in correctly classifying an action. There can be more than one action within a sliding window at the action boundaries, leading to noisy training data. Due to this reason, we do not use features extracted from frames at the action boundaries for vocabulary construction and {\sc svm} training. However, all the frames are used for testing.

The annotated dataset and the source code for the paper are available at the project page: \url{http://cvit.iiit.ac.in/research/projects/19/280/}

\subsection{Results on Different Datasets}

We follow experimental setup of Fathi {\em et. al.} \cite{fathi2011understanding} for {\sc gtea} dataset. They perform joint modelling of actions, activities and objects, on activities of three subjects and predict actions on activities of one subject. They have reported an accuracy of $47.70\%$ using their method. Table \ref{table:GTEAexperiments} summarizes our analysis of the effect of different parameters on the performance of our descriptor on the dataset.


\renewcommand{\tabcolsep}{0.25cm}
\begin{table}[t]
    \centering
    \begin{tabular}{L{2.15cm}C{3.3cm}R{1.5cm}}
        \toprule[1.5pt]
        \specialcell{\bf Method}  &\specialcell{\bf Feature} &\specialcell{\bf Accuracy} \\ \midrule


			        & trajectory & 25.36\%\\
   					& {\sc hog} & 50.17\%\\
   Uni-directional & {\sc hof} & 30.16\%\\
    Trajectory     & {\sc mbh}  & 48.69\%\\
				    & {\sc hog}+{\sc hof}+{\sc mbh}  & 50.83\%\\
    \midrule[0.5pt]
    				& trajectory & 27.09\%\\
    				& {\sc hog} & 51.25\%\\
	Bi-directional & {\sc hof} & 35.41\%\\
     Trajectory		& {\sc mbh}  & 48.87\%\\
			       & {\sc hog}+{\sc hof}+{\sc mbh}  & 54.61\%\\

    \midrule


   Affine-flow Compensation  & {\sc hog}+{\sc hof}+{\sc mbh}  & 56.87\% \\

    Camera Stabilization &  {\sc hog}+{\sc hof}+{\sc mbh}  & 57.10\% \\

    \midrule

    With 3 level Pyramid & {\sc hog}$^{Pyr}$+{\sc hof}$^{Pyr}$+{\sc mbh}  & 58.50\% \\

    \midrule







 				 &  {\sc hog}$^{Pyr}$+{\sc hof}$^{Pyr}$+{\sc mbh}  \\
 	Combined					 & and &				61.23\%\\
 	 & Kinematic + Statistical + Camera Activity &  \\

        \bottomrule[1.5pt] 
    \end{tabular}

    \caption{Effect of different parameters on the performance of our algorithm. The experiments are done on {\sc gtea} dataset. We use trajectory, {\sc hog}, {\sc hof} and {\sc mbh} features using $2K$ vocabulary size for each feature for the experiment. Accuracy reported is computed per frame.}
    \label{table:GTEAexperiments}
\end{table}

   \renewcommand{\tabcolsep}{0.3cm}
\begin{table}[t]
    \centering
    \begin{tabular}{lcc}
        \toprule[1.5pt]
        \specialcell{\bf Input}  &\specialcell{\bf PoT} &\specialcell{\bf Ours} \\ \midrule
       Raw video segment & 45.60\% & 54.61\% \\
       Stabilized video segment & 49.14\% & 61.23\% \\
        \bottomrule[1.5pt] 
    \end{tabular}
    \caption{Comparisons with Pooled Features (PoT)  \cite{ryooPOTfeature} using 3 level temporal pyramid ({\sc hog}+{\sc hof}+{\sc mbh}) on {\sc gtea} dataset.}
    \label{table:PooledFeaturesComparison}
\end{table}

We have done a comparison (Table\ref{table:PooledFeaturesComparison}) with Pooled Time Series feature \cite{ryooPOTfeature} using their released code. Our results outperforms the pooled features by a significant margin. The primary reason might be that pooled features do not seem to consider salient regions specially. We expect that there might be some merit in considering trajectory aligned pooled features.

\renewcommand{\tabcolsep}{0.07cm}
\begin{table}[t]
    \centering
    \begin{tabular}{lccc}
    \toprule[1.5pt]
    \multirow{2}{*}{ \specialcell{\bf Dataset}} & \multicolumn{3}{c}{ \specialcell{\bf Accuracy}} \\ \cline{2-4} 
        &   \specialcell{\bf Frame level} &  \specialcell{\bf Segment level} &  \specialcell{\bf Chance level} \\  \midrule
    {\sc gtea} \cite{fathi2011learning} & 61.23\% & 77.40\% & 9\%\\
    Kitchen \cite{spriggs2009temporal} & 59.74\% & 60.00\% & 3.4\%\\
    {\sc adl} \cite{pirsiavash2012detecting} & 31.40\% & 31.82\% & 4.7\%\\
    {\sc ute} \cite{lee2012discovering} & 52.62\% & 55.12\% & 4.7\%\\
    Extreme Sports & 51.20\% & 55.74\% & 5.5\% \\
    \bottomrule[1.5pt]
    \end{tabular}
    \caption{Our results for first person action recognition on different egocenric videos datasets. Sliding window based approach for classification used in our algorithm performs poorly at action boundaries. Therefore, the accuracy for segment level classification, when the action boundaries are clearly defined, comes out higher.}
    \label{table:ourBestResults}
\end{table}

We extend our experiments to other publicly available egocentric video datasets. Results on these datasets are shown in Table~\ref{table:ourBestResults}.  We follow the same experimental setup as  \cite{spriggs2009temporal} and perform frame level action recognition for `Brownie' activity for 7 subjects. Spriggs {\em et al.} \cite{spriggs2009temporal} reports an accuracy of $48.64\%$ accuracy when using first person data alone and $57.80\%$ when combined with {\sc imu} data.  We achieve $59.74\%$ accuracy using our method on egocentric video alone. 

 The {\sc adl} dataset has been used for long term activity recognition by \cite{pirsiavash2012detecting} in the past. We annotated the dataset with the short term actions and test our method on it. Similar to our experiment on {\sc gtea}, we test our model on one subject while using other for training. We achieve $31.40\%$ accuracy at frame level and $31.82\%$ at video segment level using our method. 
  Note that, {\sc adl} dataset is much larger and challenging dataset when compared to others. {\sc adl} contains actions from a diverse set of 18 activities while {\sc gtea} contains 7 activities and Kitchen dataset consider only one activity.

The {\sc ute} dataset has been used for video summarization by \cite{lee2012discovering} in the past.
 Motion blur and low image quality is fairly common in this dataset.
 For action recognition we achieve $52.62\%$ accuracy at frame level and $55.12\%$ at video segment level using our method. 

On our Extreme Sports dataset where objects and hands are not visible, our method achieves similar performance, $51.20\%$ at frame level and $55.74\%$ at segment level. Short trajectories prove to be useful for short term actions even in severe camera or head motion which is fairly common in first person videos of extreme sports.

The proposed action descriptor improves the baseline as well as the state of the art on all the five datasets tested upon (see Table \ref{table:datasets} for the details about dataset and comparison details with baseline). Figure \ref{fig:visual_res} shows some of the actions from different datasets correctly classified by our approach. Note the difference in appearance.

\subsection{Failure Analysis}

\begin{figure}[t]
    \centering
    \includegraphics[width=1.5in]{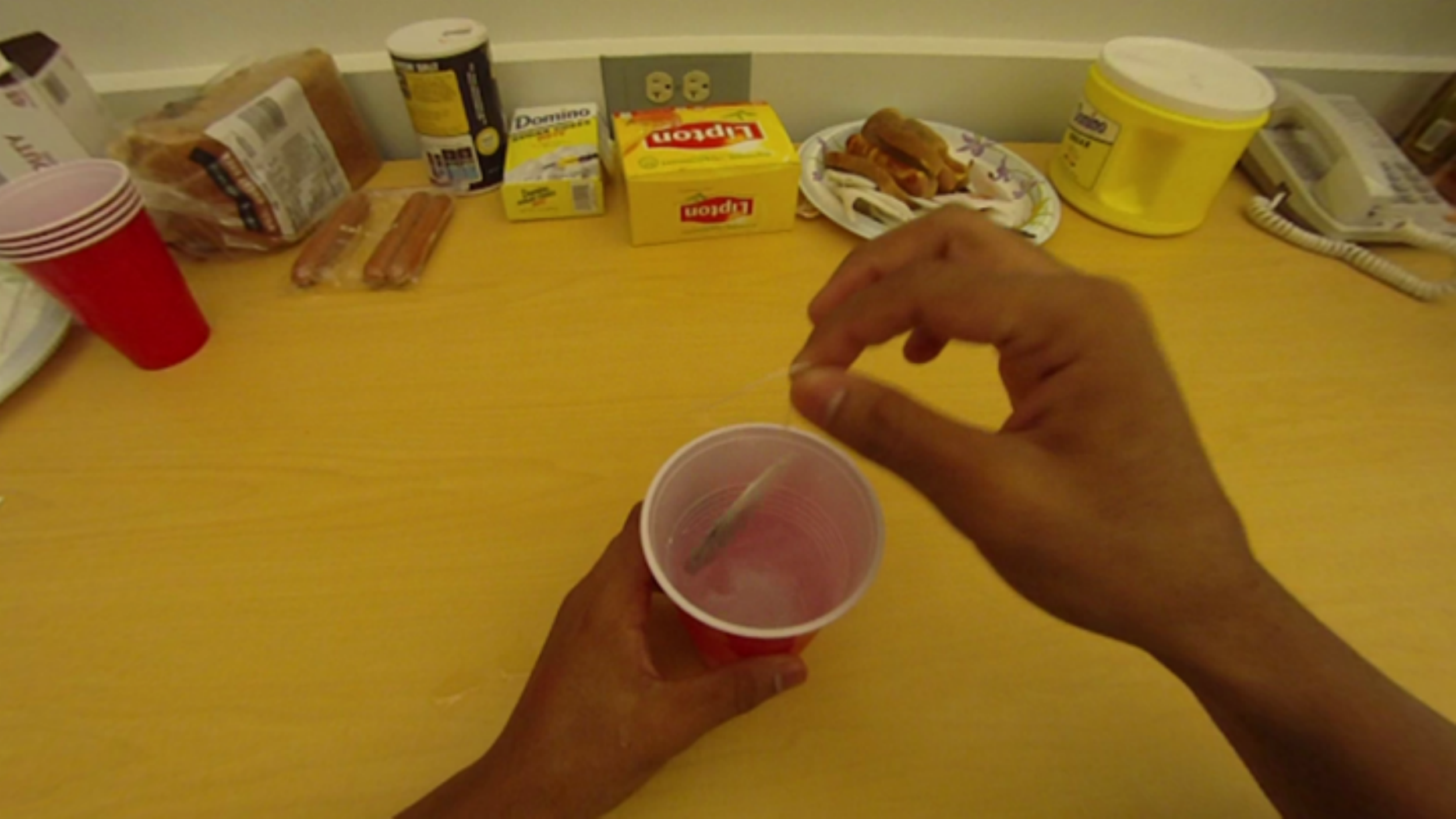}
    \includegraphics[width=1.5in]{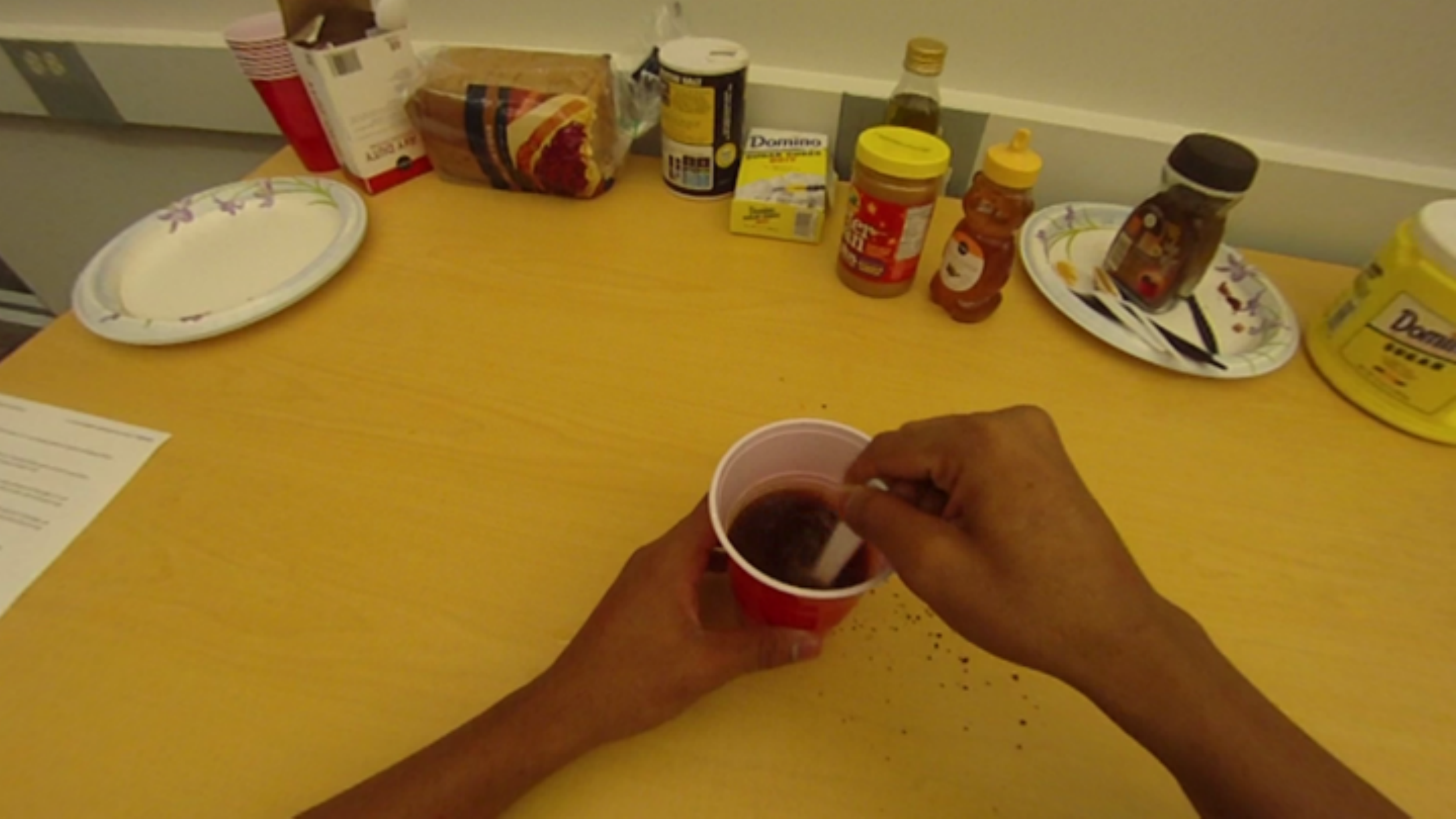} \\ \vspace{0.05cm}
    \includegraphics[width=1.5in]{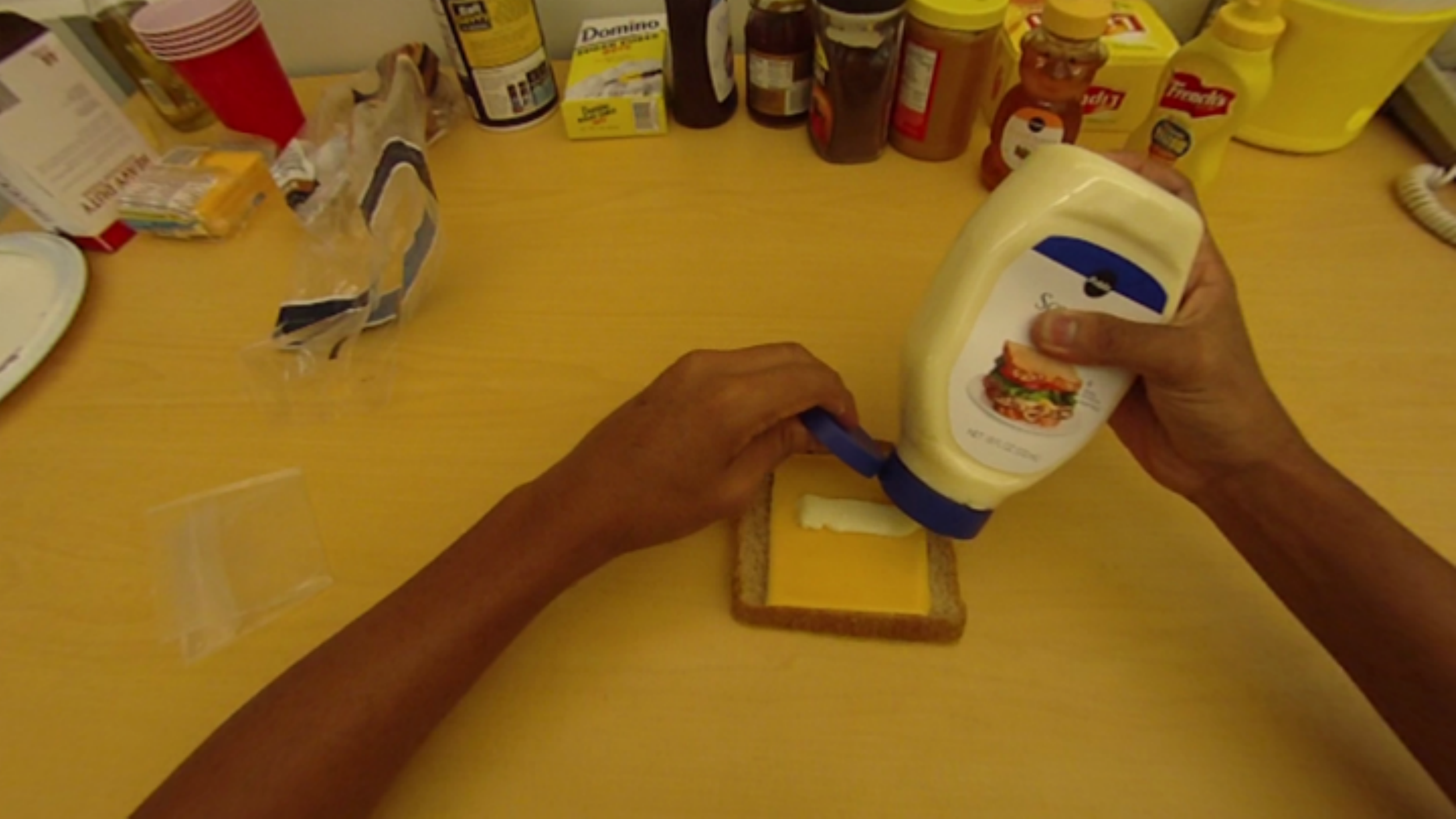}
    \includegraphics[width=1.5in]{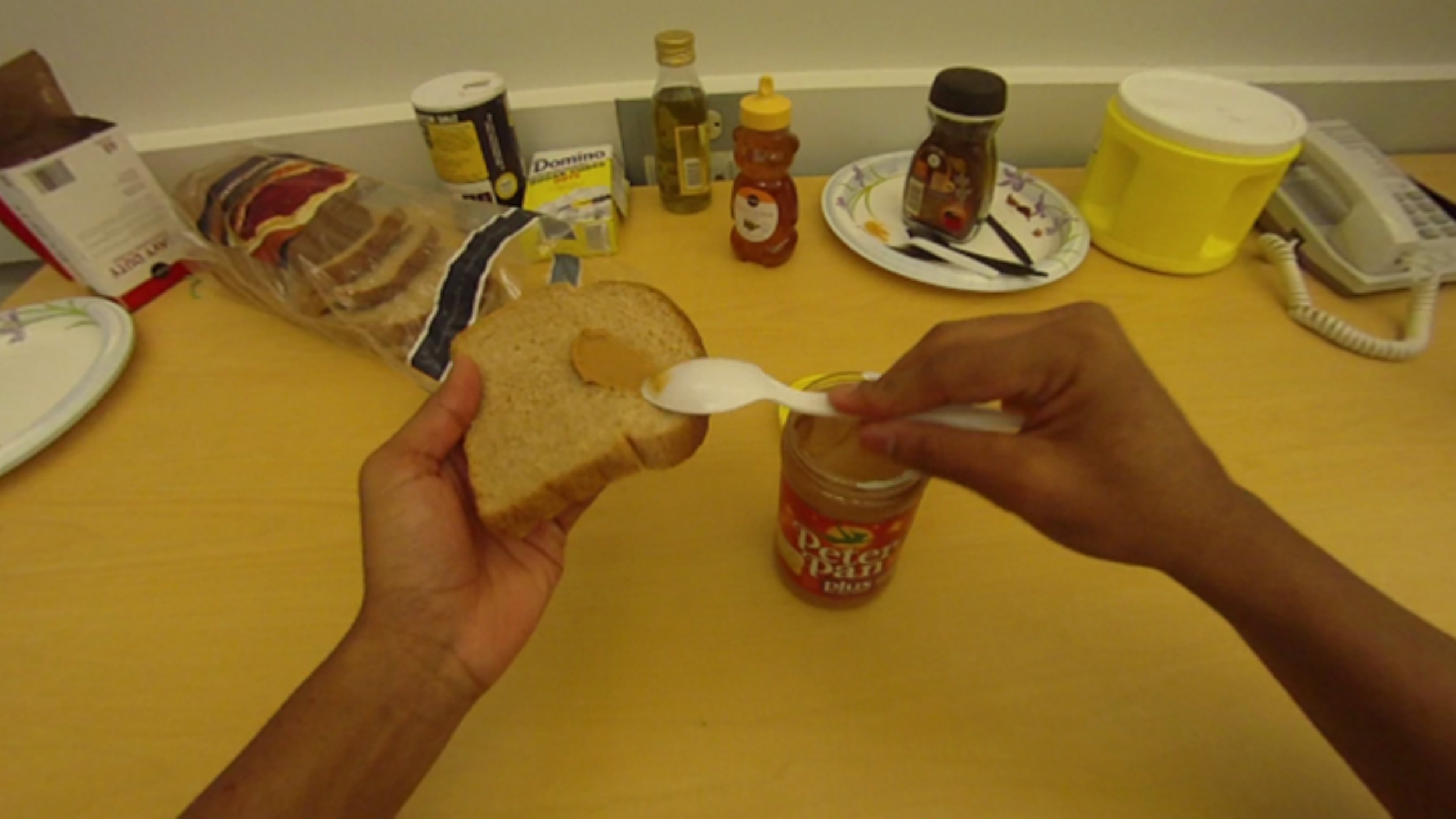}
    \caption{Some failure cases of our method. First row: `shake' classified as `stir' due to high visual and motion similarity. On the right, a similar frame with `stir' action classified correctly. Second row: `pour' classified as `spread' due to hand movement, notice the high similarity between `pouring' mayonnaise and `spreading' jam. On the right, a frame classified correctly as spread.}
    \label{fig:failure_cases}
\end{figure}

We rely on motion and appearance based cues for action recognition. While statistical and trajectory aligned features are useful for all the action classes, camera activity feature is particularly helpful with actions that has specific camera motion such as `pour', `stir' and `shake'. Though, highly discriminatory, we do see the instance when such features fail to classify correctly because of either dominant visual similarity or motion similarity or both. Figure \ref{fig:failure_cases} shows some failure cases. 

\begin{figure}[t]
    \centering
    \includegraphics[width=0.85\linewidth]{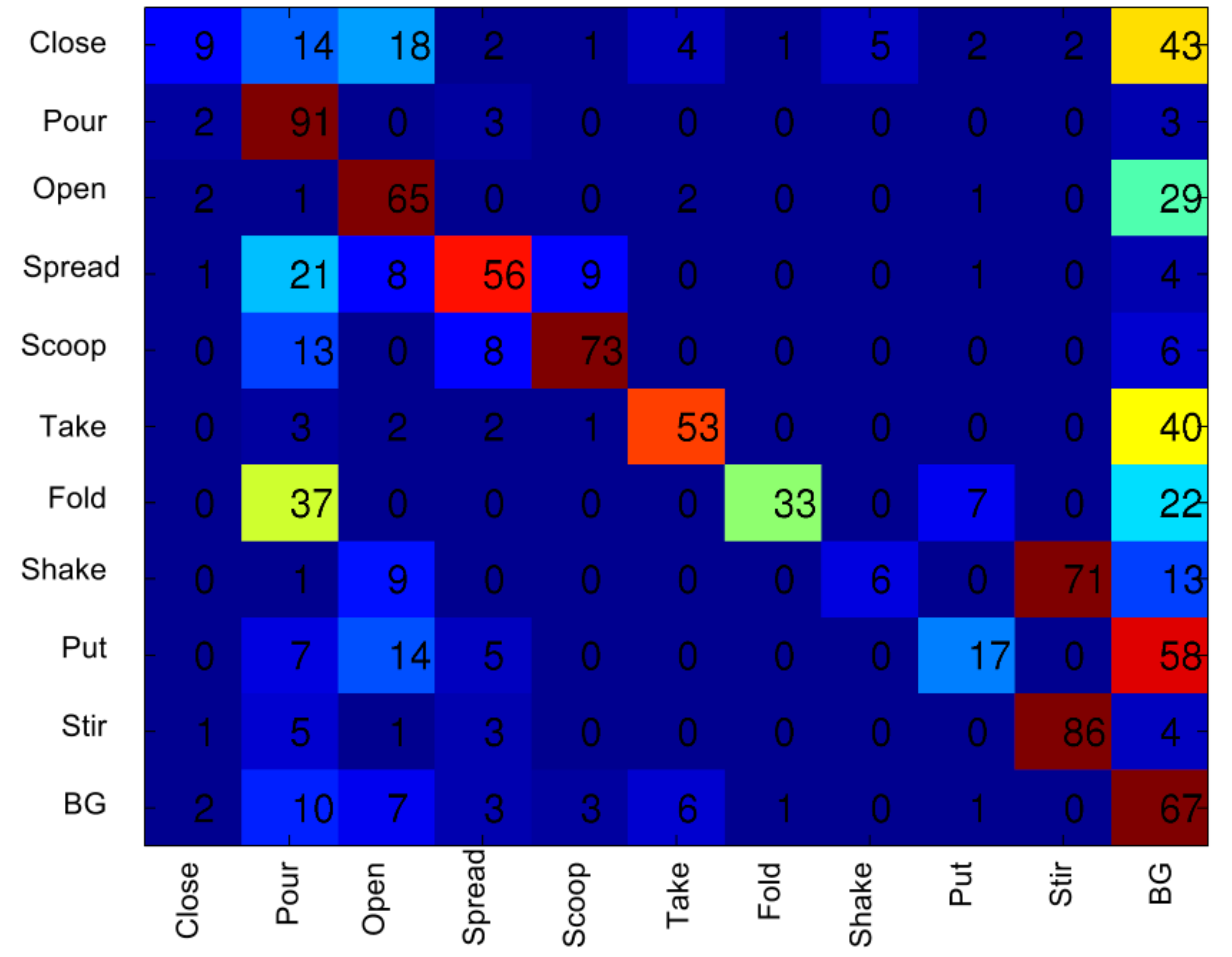}
    \caption{Confusion matrix for our method on {\sc gtea} dataset. We observe that many errors occur because action boundary is not clearly defined. `close' is commonly confused with `open' due to similarity in the nature of the action. Also, most action occurs before or after `background', hence the common confusion.
    }
    \label{figure:confusion_matrix}
\end{figure}

Figure \ref{figure:confusion_matrix} gives the confusion matrix of the proposed approach for the {\sc gtea} dataset. A large portion of observed errors occur on the action boundaries where the features from two actions merge. How to handle multiple complex actions and the action boundaries are the weak points of the proposed framework and directions for our future research. Yet some other errors arise due to limited capability of the proposed action descriptors to describe the action complexity and various ways in which the same action could have been performed. Presence of multiple actions poses another challenge. Enhancing the proposed action descriptor when the two actions are being performed jointly is another area of our future research.

\section{Conclusions}

We propose a new action descriptor for the first person action recognition from egocentric videos. In the absence of wearer's pose, the important cues for such action recognition tasks are objects present in the scene, how they are being handled and the motion of the wearer. The proposed descriptor accumulates all such cues by a novel combination of features from trajectories, {\sc hog}, {\sc hof}, {\sc mbh} as well as kinematic and statistical features. We also explore the importance of head motion and capture it using camera activity features. The proposed feature and bag of words model is able to adequately learn the representation and improves the state of the art in terms of accuracy by more than $11\%$. We validate the proposed descriptor by testing on widely varying egocentric action dataset. The performance improvement in all the datasets validates the generalizability of the proposed descriptor. Our method gives similar performance for action recognition even when handled objects or wearer's hands are not visible.

The thesis of our work and an important conceptual contribution is the observation that while objects and hands are important in first person actions, explicitly segmenting or recognising them is not necessary. It may be noted that trajectory based features can not be applied as is to egocentric actions, as shown in our baseline in Table 1. This is due to the extreme shake present in egocentric videos because of motion of wearer's head. Our second hypothesis is that for the purpose of egocentric actions, such motion can be adequately compensated using Homography alone.

Another crucial contribution is to create a bridge between first person and third person action recognition techniques. Many of the proposed features have been used in principle in problems from areas other than egocentric \cite{mihir_cvpr13, wang2011action, wang2013action}. Their use for egocentric actions now looks obvious after our experiments and findings. However, none of the prior art for egocentric actions cited in the paper have used such features.

{\footnotesize
\bibliographystyle{ieee}
\bibliography{ego_action_PRjournal}
}

\end{document}